\documentclass[onecolumn, journal]{IEEEtran}
\ifCLASSINFOpdf
\else
\fi

\usepackage{graphicx}
\usepackage{cite}
\usepackage{picinpar}
\usepackage{amsmath}
\usepackage{url}
\usepackage{colortbl}
\usepackage{soul}
\usepackage{multirow}
\usepackage{pifont}
\usepackage{color}
\usepackage{alltt}
\usepackage[hidelinks]{hyperref}
\usepackage{enumerate}
\usepackage{siunitx}
\usepackage{breakurl}
\usepackage{epstopdf}
\usepackage{pbox}

\usepackage{array}
\usepackage{threeparttable}
\usepackage{multirow}
\usepackage{algorithm}
\usepackage{algorithmic}
\usepackage{bm}
\usepackage{CJK}
\usepackage{indentfirst}
\usepackage{amsmath}
\usepackage{subfigure}
\usepackage{color}
\usepackage{booktabs}
\usepackage{amsmath,amssymb}
\usepackage{epstopdf}
\usepackage{textcomp}
\usepackage{diagbox}
\usepackage{setspace}

\usepackage{bm}
\usepackage{CJK}
\usepackage{indentfirst}
\usepackage{amsmath}
\usepackage{multirow}
\usepackage{threeparttable}
\usepackage{graphicx}
\usepackage{subfigure}
\usepackage{color}
\usepackage{booktabs}
\usepackage{epstopdf}

\DeclareMathAlphabet{\mathcal}{OMS}{cmsy}{m}{n}

\begin{document}
\title{Digital Twin for Real-time Li-ion Battery \\ State of Health Estimation with \\ Partially Discharged Cycling Data}

\author{Yan~Qin,~\IEEEmembership{Member,~IEEE,}
			Anushiya~Arunan,~\IEEEmembership{}
       	       Chau~Yuen,~\IEEEmembership{Fellow,~IEEE}        	
\thanks{This work was supported by EMA-EP011-SLEP. (Corresponding author: Chau Yuen)}
\thanks{Y. Qin, A. Arunan, and C. Yuen are with the Engineering Product Development Pillar, The Singapore University of Technology and Design, 8 Somapah Road, 487372 Singapore. (e-mail: yan.qin@ntu.edu.sg, anushiya arunan@mymail.sutd.edu.sg, yuenchau@sutd.edu.sg)}
}

\maketitle
\begin{abstract}
To meet the fairly high safety and reliability requirements in practice, the state of health (SOH) estimation of Lithium-ion batteries (LIBs), which has a close relationship with the degradation performance, has been extensively studied with the widespread applications of various electronics. The conventional SOH estimation approaches with digital twin are end-of-cycle estimation that require the completion of a full charge/discharge cycle to observe the maximum available capacity. However, under dynamic operating conditions with partially discharged data, it is impossible to sense accurate real-time SOH estimation for LIBs. To bridge this research gap, we put forward a digital twin framework to gain the capability of sensing the battery's SOH on the fly, updating the physical battery model. The proposed digital twin solution consists of three core components to enable real-time SOH estimation without requiring a complete discharge. First, to handle the variable training cycling data, the energy discrepancy-aware cycling synchronization is proposed to align cycling data with guaranteeing the same data structure. Second, to explore the temporal importance of different training sampling times, a time-attention SOH estimation model is developed with data encoding to capture the degradation behavior over cycles, excluding adverse influences of unimportant samples. Finally, for online implementation, a similarity analysis-based data reconstruction has been put forward to provide real-time SOH estimation without requiring a full discharge cycle. Through a series of results conducted on a widely used benchmark, the proposed method yields the real-time SOH estimation with errors less than 1$\%$ for most sampling times in ongoing cycles.
\end{abstract}

\IEEEpeerreviewmaketitle

\section{Introduction}
\IEEEPARstart{T}{he} global climate change and energy crisis have stimulated renewable energy development and booming demands for Lithium-ion batteries (LIBs). LIBs are recyclable electrical energy storage components with high energy density, low cost, and high portability \cite{Ref4}. The whole world has witnessed the vital role of LIBs in the incoming energy revolution, and the major economies have put forward a series of policies and given extensive financial support to spur the development of LIBs and the relevant industries, such as electric vehicles (EVs) \cite{RefEditor1} and smart grid \cite{Ref1}. For this reason, ubiquitous LIBs have been observed, from electric transportation to energy storage to replace traditional fossil fuels, specifically in the field of the Industrial Internet of Things (IIoT). IIoT has fundamentally transformed industrial predictive management by interconnecting and leveraging data for their benefits \cite{IIoT}, offering a digitalization solution to avoid the abrupt breakdown of critical assets. LIBs have been an essential part of this evolution and are widely deployed in remote IIoT devices or placed on moving assets that cannot be connected to the electricity grid. Besides, LIB has received increasing attention in grid-level energy storage, balancing power generation and utilization \cite{IIoT1}. Considering that the healthy performance of LIBs has prominent effects on their powered devices, it is a compelling research topic to timely evaluate the health status of LIBs throughout their lifespan.

Nowadays, the prospect of the digital twin has attracted extensive attention to fully understand the degradation behavior and avoid anomalous events in the predictive maintenance field \cite{DigitalTwin1}. The superiority of the digital twin paradigm is to mirror the complicated physical systems with sensing data into an interpretable and straightforward digitalization world \cite{DigitalTwin2}. The digital twin has attracted much attention from the research field of predictive maintenance since 2012 \cite{DigitalTwinNASA}, when the National Aeronautics and Space Administration raised higher requirements for future vehicles. Khan et al. \cite{DTKhan} presented the general requirements of a successful digital twin for autonomous maintenance. Ren et al. highlighted the potential of the digital twin for predictive maintenance with the example of diesel locomotives \cite{DTRen}. Although the digital twin is in its infancy for predictive maintenance, it is becoming a research hotspot. In terms of battery health evaluation, the digital twin brings a promising way for accurate sensing of the physical battery in the virtual world \cite{DigitalTwin3} through accessible measurements, such as charge/discharge voltage, surface temperature, and charge/discharge current, etc. Battery state of health (SOH) estimation allows for the determination of available capacity for used batteries, enabling the proactive replacement of exhausted batteries in advance.

The early digital twin models for SOH estimation relate historical cycling data or health indicators with labeled maximum available capacity (MAC) using various machine learning models. Before the invention of deep learning approaches, statistical machine learning models, such as Gaussian process regression, support vector machine, and random forest regression, have been popular. For instance, Liu et al. \cite{HI1} constructed health indicators from charging measurements and then adopted Gaussian process regression to predict the capacity. Li et al. \cite{HI2} adopted the random forest regression to regress the measurements on historical capacities to figure out the incoming capacities. However, limited by the short-length time series processing ability, the statistical machine learning approaches fail to extract useful information from the battery cycling data completely. Since 2016, the long short-term memory network (LSTM) \cite{LSTM1997} has become one of the most prevalent advanced machine learning techniques for SOH estimation, which exhibits superiority in memorizing crucial information from long sequence time series.

Motivated by the temporal learning ability of LSTM, You et al.\cite{RealSOH2} designed a recurrent neural network-based digital twin for SOH estimation to handle sequential data. Zhang et al. \cite{LSTM2018} incorporated the dropout mechanism into the LSTM-based digital twin for SOH estimation to resolve the problem of over-fitting during the training phase. To deal with the regeneration phenomenon of LIBs, Ma et al. \cite{Ref19} proposed a hybrid digital twin by combining convolutional neural network and LSTM to yield MAC with a sliding window technology. Benefiting from the fine-scaled decomposition of raw measurements with empirical mode decomposition, Liu et al. \cite{RefTIE} decomposed the battery capacity data into a series of components corresponding to different frequencies, and these components serve as the input of the LSTM-based digital twin for SOH estimation to regress on MAC. It is worth pointing out that the aforementioned approaches require the entire cycling data as inputs for both offline training and online application. In other words, a battery has to be cycled from fully charged status to fully discharged status during each cycle. In contrast, the partially charging/discharging profiles-based solution offers a promising alternative. Richardson et al. \cite{Richardson} constructed the input data with the fixed-length charging sequence from the given starting voltages. Liu et al. \cite{Liu} manufactured the feature by computing the voltage drop from a specific starting voltage to the point with a fixed charging length. Although data requirements can be satisfied during offline modeling, it is high requirement when these developed models are deployed for practical demands. Therefore, the current digital twins miss the real-time SOH estimation ability during online applications with partially discharged cycling data.

Accurate real-time SOH estimation with digital twin promotes high-efficient energy management through timely informing battery health status to avoid the strict requirement of a complete discharge cycle. To facilitate real-time SOH estimation, several research gaps and challenges are further highlighted regarding the complex degradation data characteristics. The first critical challenge is the variable cycling data structure, which is caused naturally by the degradation procedure. Second, the temporal importance of samples, indicating the importance of different discharge voltage areas, has not been discussed yet in the current literature. For online applications, the challenge is reconstructing the partially discharged data to yield SOH estimation with the proposed offline model. In this regard, more detailed research gaps analysis of the digital twin for real-time SOH estimation with partially discharged cycling data are given as follows:

\begin{itemize}
\item \textbf{Variable Cycling Data Structure}: A standard data structure is required to make full use of the strength of advanced machine learning. However, as MAC degrades with the increase of cycle, the decrease in discharging time leads to variable cycling data in the physical battery system. The inappropriate processing of variable cycling data probably leads to information loss or distortion in the digital world, undermining the quality of the digital twin-based SOH estimation model. Therefore, finding a proper subspace to align the variable cycling data for offline modeling and online application is crucial yet unsolved.
\item \textbf{The Different Temporal Importance}: The conventional approaches assume that different samples within a cycle are equally contributed to capacity degradation. In fact, data similarity at specific sampling times over cycles is high, meaning that these samples are less important for capacity degradation over cycles. In contrast, assigning more weights to crucial sampling times is reasonable. Therefore, temporal importance analysis helps to exclude unimportant samples during modeling and ensures the high estimation accuracy of digital twin-based SOH estimation at important sampling times.
\item \textbf{Real-time Cycling Data Reconstruction}: A practice-oriented digital twin expects to yield SOH estimation on the fly with real-time incoming data. For the online application, only data up to the present time are available. As such, the unknown future data challenge the real-time SOH estimation, as the MAC estimation is an accumulative behavior of the entire cycling data. One feasible solution is to reconstruct the missing future data.
\end{itemize}

This article proposes a digital twin-based real-time SOH estimation model to address the above-concerned challenges, consisting of three components. First, to handle the variable cycling data, cycling synchronization is approached through the proposed energy discrepancy-aware time warping to align cycling data. The alignment ensures that cycling data have the same structure by finding the proper subspace to maintain the temporal and spatial consistency between cycles to avoid information loss or distortion. Second, to evaluate the importance of different sampling times, the synchronized cycling data serve as the inputs of time-attention LSTM to regress on MAC to establish the regression model. Third, to provide real-time SOH at any moment of incoming cycles, future data reconstruction is explored according to the known data behavior. By matching the most similar time series from training data, similarity metrics are used to ensure the quality of the reconstructed future data. The performance of real-time SOH estimation is verified via an experimental dataset. Overall, a real-time SOH estimation framework with a digital twin has been achieved for the first time, providing a promising solution to estimate MAC on the fly. The major contributions of this work are summarized as follows:

\begin{itemize}
\item A variable cycling data synchronization approach has been proposed using energy discrepancy-aware time warping, aligning the variable cycling data into the same structure to facilitate the digital twin modeling.
\item A time-attention SOH estimation model has been constructed with the importance analysis of sequential samplings, enhancing the estimation accuracy with fewer samples compared to the traditional methods.
\item A future data reconstruction strategy for real-time SOH estimation has been proposed with data matching and construction strategy, facilitating the real-time digital twin of SOH estimation by evaluating the trajectory behavior.
\end{itemize}

The remaining parts of this article are structured as follows: Section II investigates the data structure of LIBs and formulates the mathematical description of the real-time SOH estimation. The details of the proposed method are explained in Section III. Further, the performances of the proposed method are demonstrated and compared through an experimental case in Section IV. We conclude this work and highlight future research in the last section.

\section{Data Description and Problem Formulation}\label{Section2}
This section describes the variable cycling data structure and mathematically formulates the real-time SOH estimation problem for LIBs. Besides, the differences between the current digital twin for SOH estimation and the proposed digital twin for real-time SOH estimation are clarified.

\begin{figure}[!ht]
\centering
\subfigure[]
{
\begin{minipage}[t]{0.45\linewidth}
\centering
\includegraphics[width=8cm]{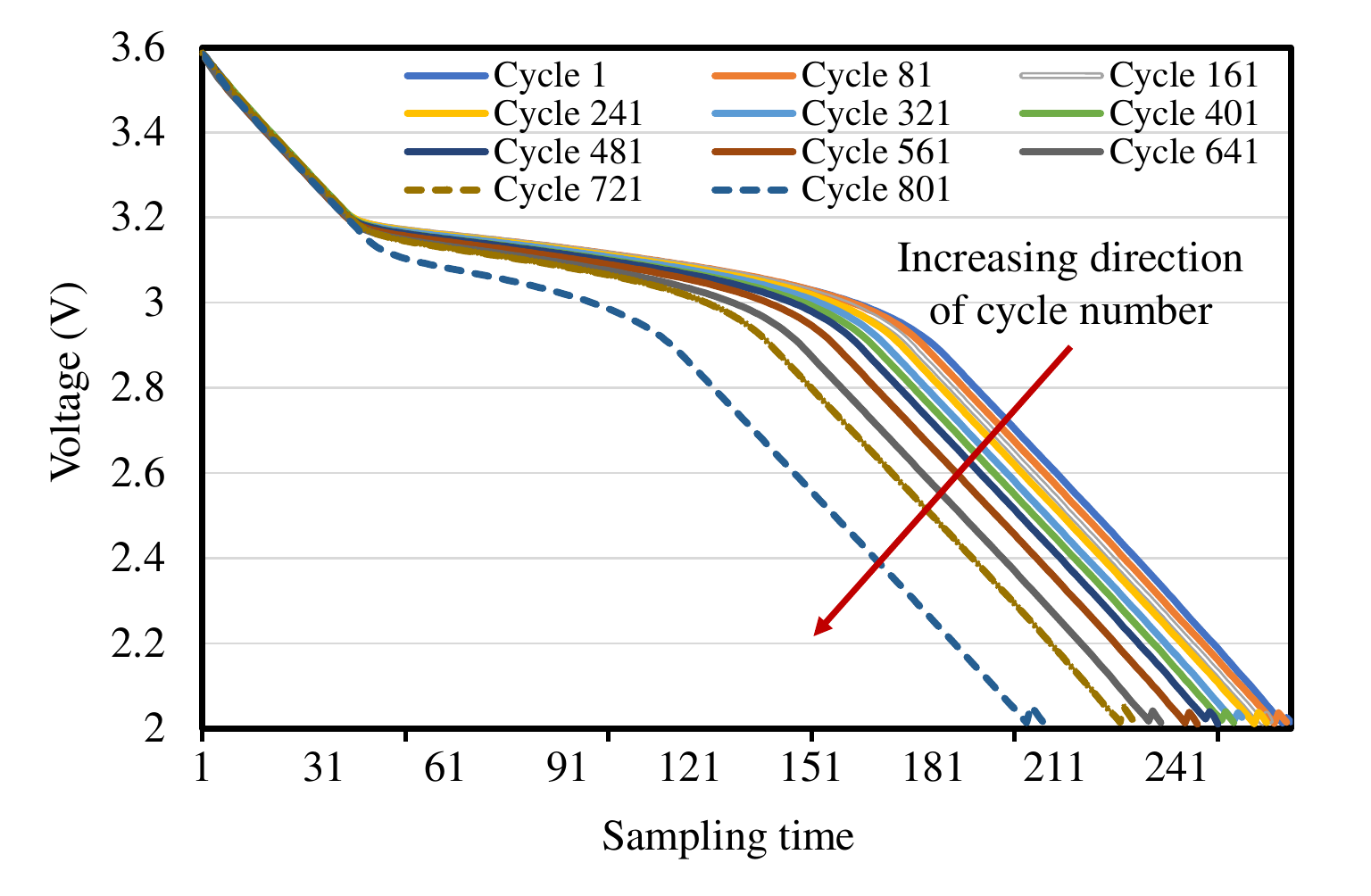}
\end{minipage}
}
\subfigure[]
{
\begin{minipage}[t]{0.45\linewidth}
\centering
\includegraphics[width=8cm]{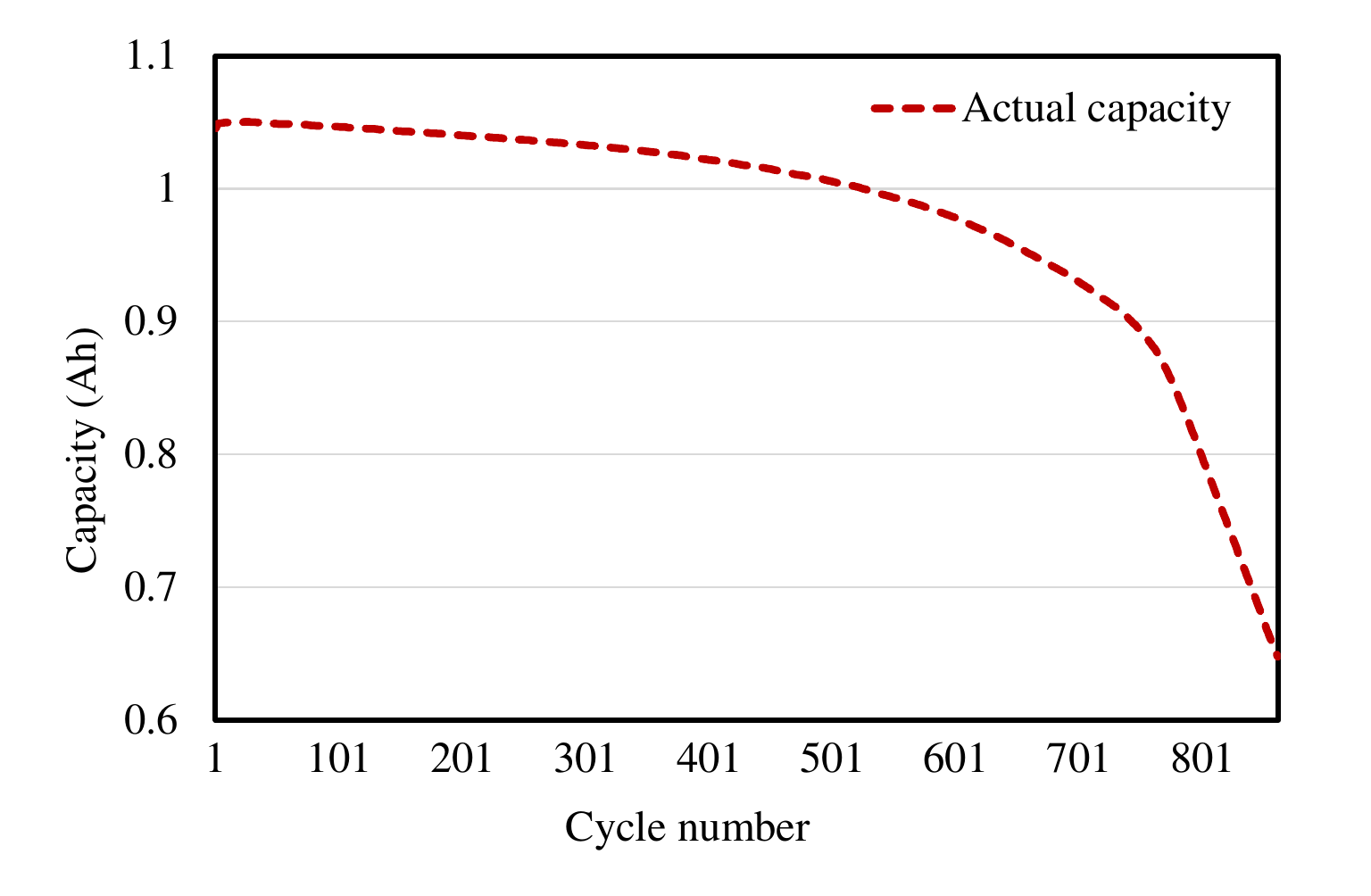}
\end{minipage}
}
\caption{Visualization of Battery 22 from the Massachusetts Institute of Technology dataset \cite{RefNature} regarding (a) discharging voltage curves with the cycle interval of eighty and (b) capacity fading over cycles.}
\label{FIG1}
\end{figure}

\subsection{Variable Cycling Data Structure with Cyclic Aging}\label{Section2.1}
LIBs follow a repetitive but non-consistent manner to absorb and release electrical energy through continuous charging and discharging. Specifically, with the external power source, Li-ions are pushed from the anode side into the cathode. As such, the battery voltage increases gradually to its maximum and then keeps constant until the charging current is below a certain threshold, e.g., 20mA. During the charging procedure, Li-ions stored in the anode move back to the cathode when an external load is connected across the battery terminals \cite{BatteryCycling}. Since the discharging procedure is cycle-varying and closely relates to the battery-powered device in real-time, this research mainly investigates the cycling data during discharging. The term cycling data indicates the time series with multiple variables corresponding to a specific discharging cycle. On the basis of this, the complete battery degradation procedure consists of numerous cycles. Assuming all cycling data are fully charged and discharged, it observes a phenomenon of variable length among cycling data, as illustrated in Fig. 1 with a LIB cell provided by the Massachusetts Institute of Technology dataset \cite{RefNature}.

\begin{figure}[!htb]
\centering
\includegraphics[scale=0.65]{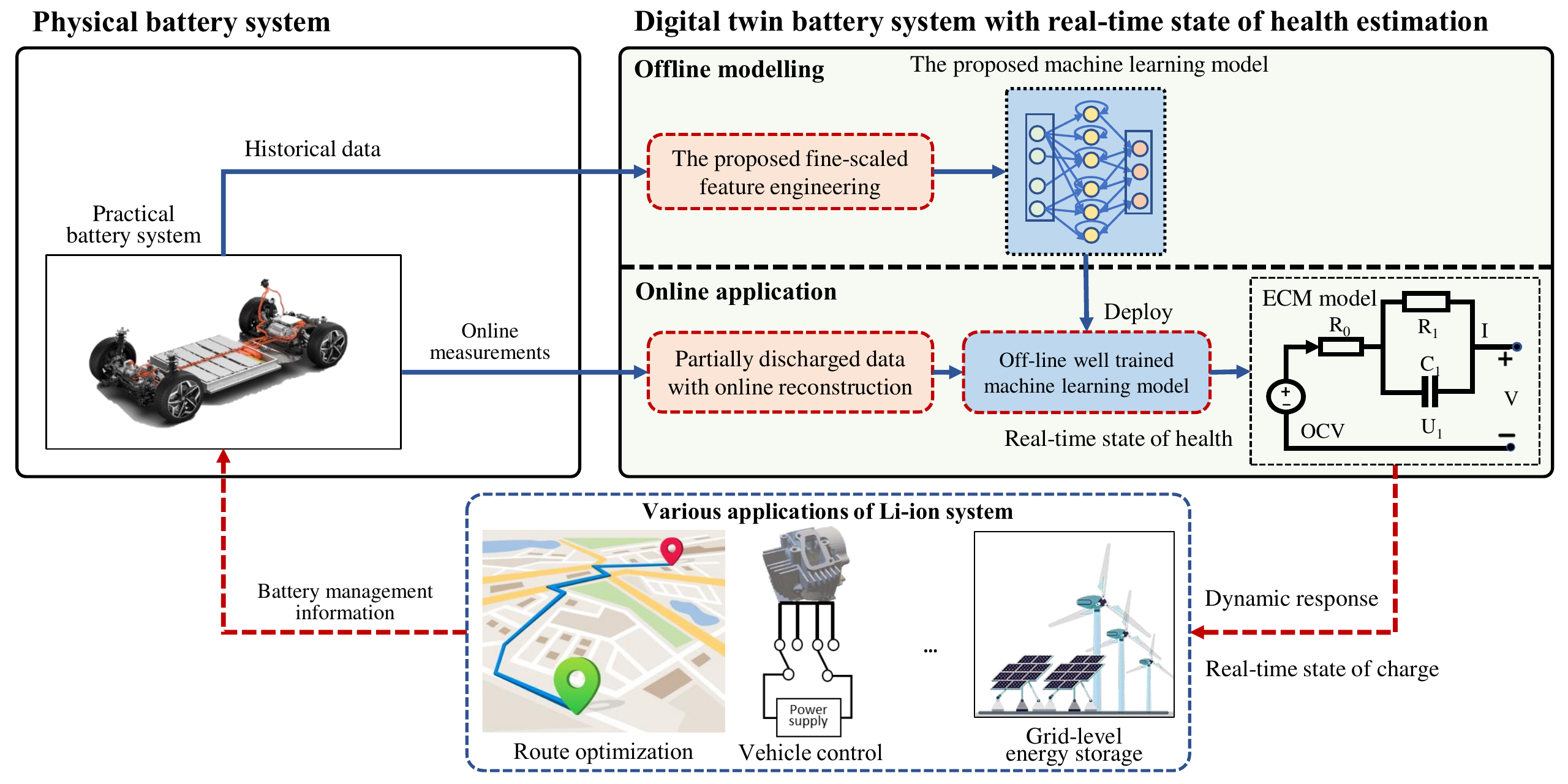}
\caption{The overall structure and information flow of the proposed digital twin for real-time SOH estimation.}
\label{FIG2}
\end{figure}

Given that $J$ variables are measured for a discharging cycle $c$, its cycling data can be described as $\mathbf X_c \in \mathbb{R}^{K_c \times J}$, where $K_c$ is the length of the $c^{th}$ cycle. It is worth noting that the collection of cycling data $\bar{\mathbf X} \in \mathbb{R}^{K_c \times J \times C}$ is not a standard three-dimensional matrix due to the variable cycle length, where $\bar{\mathbf X}(:,:,c)$ is equal to $\mathbf X_c$ and $C$ is the total number of discharging cycles.

\subsection{Digital Twin for Real-time SOH Estimation with Partially Discharged Cycling Data}\label{Section2.2}
Currently, the most popular way to evaluate the health status of a LIB is the value of MAC, which is commonly measured at the end of the discharge cycle. For clarity, we denote this kind of capacity as $H_c$ for the $c^{th}$ cycle, and ground truth of MAC of a battery are measured as $\mathbf H = [H_{1}, H_{2}, \dots, H_{c}, \dots, H_{C}]$.

Although end-of-cycle capacity is a useful metric to evaluate the health state, it is only available at the end of the discharging cycle, as the complete time series are required for modeling. To trace real-time SOH estimation, we aim to predict the capacity with partial information $\mathbf X_c(1:k)=[\mathbf x_c(1), \mathbf x_c(2), \cdots, \mathbf x_c(k)]$ up to current time step $k$ of the Cycle $c$. Afterward, we continuously update the estimations with the arrival of new samples. The real-time SOH estimation attempts to predict capacity at any sampling time, which is defined as follows:
\begin{equation}
H_c(k) = \Psi (\mathbf X_c(1:k)) \quad k \in [1, K_c]
\label{Eq1}
\end{equation}
where $H_c(k)$ is the estimated capacity at time step $k$ in the $c^{th}$ cycle, and $\Psi (\cdot)$ is the function to yield real-time estimation.

Fig. \ref{FIG2} demonstrates the LIB digital twin framework with real-time SOH estimation using the proposed approach. The traditional SOH estimation model is developed using historical data and applies the well-trained model for the incoming data. Although this kind of SOH estimation model is easy to implement as no expert knowledge or first-principles mechanism is required, it only yields estimation at the end of the discharging cycle. However, the missing real-time SOH estimation significantly impedes the accurate state estimation of the battery system, which can be reached via an equivalent circuit model (ECM). ECM is the virtual mapping of the physical battery in the digital world, and it can be derived through physics-informed technology to monitor and control LIBs. Detailed information about the ECM model is omitted here for brevity, but references \cite{ECM}, \cite{ECMMPC} are recommended for a better understanding of ECM. Timely updating of SOH is essential to gain an accurate ECM model \cite{ECM}. As shown in Fig. \ref{FIG2}, the proposed method attempts to attain the desired real-time SOH estimation ability with the interactive mechanism of offline model and real-time measurements without requiring the full discharge cycle. Through offering real-time SOH information, the ECM model is enhanced and outputs reliable battery health status for the associated services, like vehicle control. Specifically, model predictive control (MPC) has been widely reported in the driving control of EVs because of the benefits of robustness for uncertainty and disturbance and the capability of dealing with time delay \cite{MPCControl}. MPC-based driving control is observed in the heavy dependence on accurate values of SOH and state-of-charge (SOC). As such, the discrepancy between the virtual model and the physical system may lead to a discount on control performance and energy efficiency.

\begin{figure}[!htb]
\centering
\includegraphics[scale=0.34]{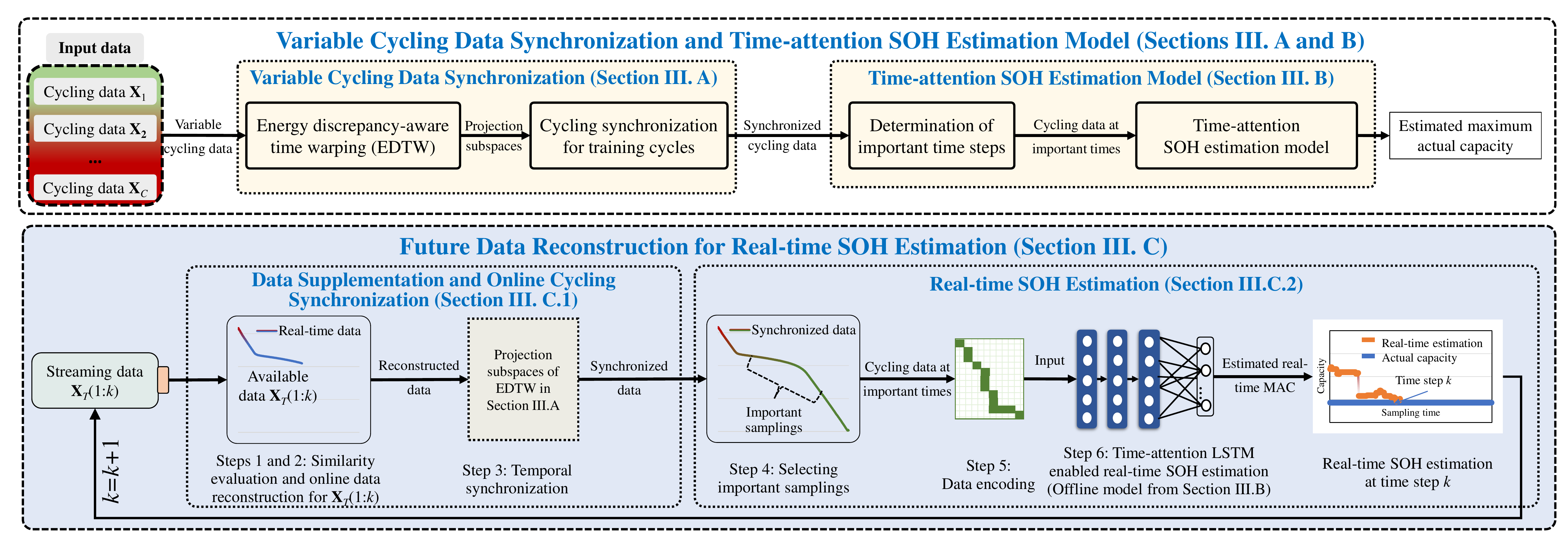}
\caption{Overview of the proposed real-time SOH estimation empowered by the digital twin framework.}
\label{FIG3}
\end{figure}

\section{Proposed Digital Twin assisted Real-time State of Health Estimation}\label{Section3}
Fig. \ref{FIG3} displays the overall framework of the proposed approach, which consists of three parts to make real-time estimation with digital twin function well. The first part offers offline model training, in which variable cycling data are synchronized. Next, a time-attention LSTM model is constructed to serve as the prediction model. The third part enables real-time estimation through data matching and reconstruction of within-cycle future trajectory, yielding the estimation of MAC at each sampling time rather than at the end of a cycle.

\subsection{Variable Cycling Data Synchronization}\label{Section3.1}
\subsubsection{Energy Discrepancy-aware Time Warping (EDTW)}
Although few studies are available for cycling synchronization of battery data, time-series synchronization has been an active topic in many research fields, such as computer vision \cite{CVDTW}, batch process \cite{BatchDTW}. Typical time-series synchronization approaches are briefly revisited here for better understanding. Dynamic time wrapping (DTW) has been historically a fundamental method to synchronize time series with variable length, which finds the optimal path where two aligned time series have the minimum distance. However, DTW overlooks the temporal correlation of time series. An extension of DTW, called canonical time warping (CTW) \cite{CTW},  was put forward to acquire spatial-temporal synchronization of two motion trajectories. Zhou et al. further \cite{GTW} extended CTW into a more general framework by aligning multiple time series.

\begin{algorithm}[!ht]
\scriptsize
\caption{The iterative solution of EDTW}
\label{alg1}
\begin{algorithmic}[1]
\STATE Input:  $\mathbf X_s$ and $\mathbf X_t$
\STATE Output: $\mathbf W_r$, $\mathbf W_t$, $\mathbf V_r$, and $\mathbf V_t$
\STATE Initialize $\mathbf V_r$ = $\mathbf I_{K_r}$, $\mathbf V_t$ = $\mathbf I_{K_t}$
\FOR{$i = 1, 2, ..., N$}
\STATE Get the sequences for alignment as $\hat {\mathbf V}_r^T = \mathbf V_r^T \mathbf X_r$ and $\hat {\mathbf V}_t^T = \mathbf V_t^T \mathbf X_t$
\STATE Obtain the matrixes $\mathbf W_s$ and $\mathbf W_t$ using the traditional DTW algorithm to synchornize $\hat {\mathbf V}_r^T$ and $\hat {\mathbf V}_t^T$;
\STATE Calculate the new $\mathbf V_r$, $\mathbf V_t$ corresponding to the maximum eigenvalue with the following eigenvalue decomposition problem:
\begin{equation}
\begin{aligned}
\mathbf X_r \mathbf W_r^T \mathbf W_r \mathbf X_r^T \mathbf V_r = \mathbf X_r^T \mathbf W_r^T \mathbf W_t \mathbf X_t^T \mathbf V_t \Delta_1   \\
\mathbf X_t \mathbf W_t^T \mathbf W_t \mathbf X_r^T \mathbf V_t = \mathbf X_t^T \mathbf W_t^T \mathbf W_r \mathbf X_r^T \mathbf V_r \Delta_2   \\
\end{aligned}
\label{Eq2}
\end{equation}
\STATE With $\mathbf W_r$, $\mathbf W_t$, $\mathbf V_r$, and $\mathbf V_t$, obtain the objective given in Eq. (4);
\IF {$\Pi$ converges to a predefined small value, e.g., 0.01.}
\STATE $i=i+1$ and repeat above procedures until the convergence condition is meet or the maximum iteration steps is reached
\ENDIF
\ENDFOR
\end{algorithmic}
\end{algorithm}

\begin{figure}[!htb]
\centering
\includegraphics[scale=0.8]{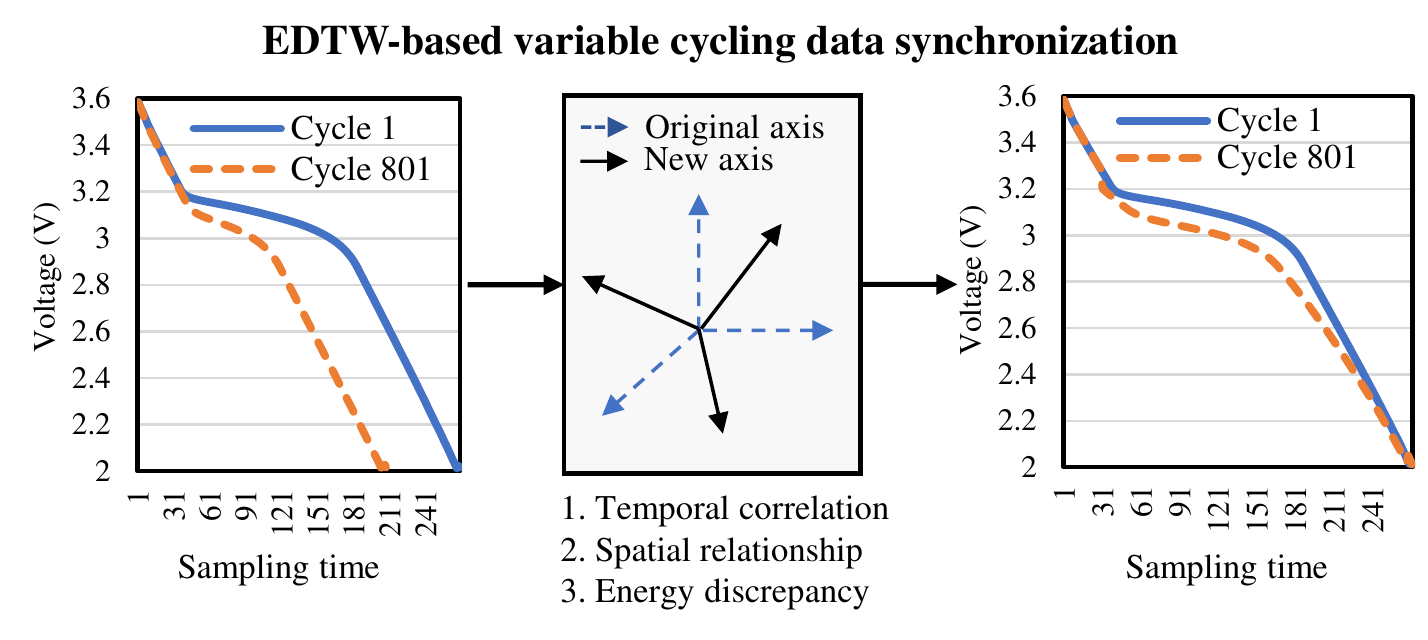}
\caption{The illustration of cycling synchronization using the proposed EDTW to align variable cycling data, where two random cycles from Battery 22 of Massachusetts Institute of Technology dataset \cite{RefNature} are used for demonstration.}
\label{FIG4}
\end{figure}

Since the objective function of CTW minimizes the difference between two time series, it is improper to be directly applied for battery cycling data since the discrepancy between cycles will be erased. However, the variable length of cycling data is an important indicator of performance degradation, which leads to energy discrepancy among cycling data. It is reasonable to consider energy discrepancy during cycling synchronization to ensure the final estimation accuracy. As shown in Fig. \ref{FIG4}, synchronization of battery data with variable lengths should take the following concerns into account:
\begin{itemize}
\item \textit{Temporal correlation}: Temporal alignment of discharging procedure behaves similar shapes across different cycles;
\item \textit{Spatial relationship}: The shape of the synchronized cycling data are expected to keep similar with the cycling data before synchronization. For instance, discharging voltage varies from high to low for each cycle;
\item \textit{Energy discrepancy}: The energy of a cycle is expected to be the same before and after synchronization.
\end{itemize}

Given two cycling trajectories $\mathbf X_r \in \mathbb{R}^{K_{r} \times J}$ and $\mathbf X_t \in \mathbb{R}^{K_{t} \times J}$, EDTW handles the desired purposes to design cycling synchronization to minimize the following objective function:
\begin{equation}
\begin{aligned}
min~\Pi & = \|\mathbf{V}_{r}^{\mathrm{T}} \mathbf{X}_r \mathbf{W}_{r}^{\mathrm{T}}-\mathbf{V}_{t}^{\mathrm{T}} \mathbf{X}_t \mathbf{W}_{t}^{\mathrm{T}}\|^{2}  \\
& + (\| \mathbf{V}_{r}^{\mathrm{T}}\mathbf{X}_r\mathbf{W}_{r}^{\mathrm{T}}\mathbf{W}_{r}\mathbf X_r^{\mathrm{T}} \mathbf{V}_{r} \|^2- \| \mathbf X_r^{\mathrm{T}} \mathbf X_r \|^2) \\
& + (\| \mathbf{V}_{t}^{\mathrm{T}} \mathbf{X}_t \mathbf{W}_{t}^{\mathrm{T}}\mathbf{W}_{t}\mathbf X_t^{\mathrm{T}} \mathbf{V}_{t} \|^2- \| \mathbf X_t^{\mathrm{T}} \mathbf X_t \|^2)
\end{aligned}
\label{Eq3}
\end{equation}
subject to,
\begin{equation}
\begin{aligned}
& \mathbf{X}_r \mathbf{W}_{r}^{\mathrm{T}} \mathbf{1}_{m} = \mathbf{0}_{K_r}, \mathbf{X}_t \mathbf{W}_{t}^{\mathrm{T}} \mathbf{1}_{m}= \mathbf{0}_{K_t} \\
& \mathbf{V}_{r}^{\mathrm{T}} \mathbf{X}_r \mathbf{W}_{r}^{\mathrm{T}} \mathbf{W}_{r} \mathbf{X}_r^{\mathrm{T}} \mathbf{V}_{r} = \mathbf{V}_{t}^{\mathrm{T}} \mathbf{X}_t \mathbf{W}_{t}^{\mathrm{T}} \mathbf{W}_{t} \mathbf{X}_t^{\mathrm{T}} \mathbf{V}_{t} = \mathbf{I}_{M} \\
& \mathbf{V}_{r}^{\mathrm{T}} \mathbf{X}_r \mathbf{W}_r^{\mathrm{T}} \mathbf{W}_t^{\mathrm{T}} \mathbf{X}_t^{\mathrm{T}} \mathbf{V}_{t} = \mathbf{\Lambda}
\end{aligned}
\label{Eq4}
\end{equation}
where $\mathbf X_r$ is the reference cycling trajectory and  $\mathbf X_t$ indicates the trajectory to be aligned; $\mathbf V_r \in \mathbb{R}^{K_r \times M}$ and  $\mathbf V_t \in \mathbb{R}^{K_t \times M}$ determine the spatial warping by projecting the sequences into the same coordinate system corresponding to $\mathbf X_r$ and $\mathbf X_t$, where $M \leq \min \left(K_{r}, K_{t}\right)$; $\mathbf W_r$ and $\mathbf W_t$ warp the reference and target signals in time to achieve optimum temporal alignment; $m$ is the number of steps required to align cycling times; $\mathbf{\Lambda}$ is a diagonal matrix.

The first constraint in Eq. (\ref{Eq4}) encodes the alignment path to follow continuity and monotonicity. The second constraint ensures the decomposed features in $\mathbf{V}_{r}^{\mathrm{T}} \mathbf{X}_r \mathbf{W}_{r}^{\mathrm{T}}$ are orthogonal to each other. The same operation is conducted for $\mathbf{V}_{t}^{\mathrm{T}} \mathbf{X}_t \mathbf{W}_{t}^{\mathrm{T}}$ in the second constraint. The third constraint maximizes the correlation between a pair of decomposed features from $\mathbf{X}_{r}$ and $\mathbf{X}_{t}$, while keeping them uncorrelated with other features.

The Lagrange multiplier is employed to handle the constraints, and the objective function is rewritten as below:

\begin{equation}
\begin{aligned}
min~\Pi & = \|\mathbf{V}_{r}^{\mathrm{T}} \mathbf{X}_r \mathbf{W}_{r}^{\mathrm{T}}-\mathbf{V}_{t}^{\mathrm{T}} \mathbf{X}_t \mathbf{W}_{t}^{\mathrm{T}}\|^{2}  \\
& + (\mathbf{1}_{M}^{\mathrm{T}} \mathbf{V}_{r}^{\mathrm{T}}\mathbf{X}_r\mathbf{W}_{r}^{\mathrm{T}}\mathbf{W}_{r}\mathbf X_r^{\mathrm{T}} \mathbf{V}_{r}\mathbf{1}_{M} - \mathbf{1}_{K_r}^{\mathrm{T}} \mathbf X_r^{\mathrm{T}} \mathbf X_r \mathbf{1}_{K_r}) \\
& + (\mathbf{1}_{M}^{\mathrm{T}} \mathbf{V}_{t}^{\mathrm{T}} \mathbf{X}_t \mathbf{W}_{t}^{\mathrm{T}}\mathbf{W}_{t}\mathbf X_t^{\mathrm{T}} \mathbf{V}_{t} \mathbf{1}_{M} - \mathbf{1}_{K_t}^{\mathrm{T}} \mathbf X_t^{\mathrm{T}} \mathbf X_t \mathbf{1}_{K_t}) \\
& + \lambda(\mathbf{V}_{r}^{\mathrm{T}} \mathbf{X}_r \mathbf{W}_{r}^{\mathrm{T}}\mathbf{W}_{r} \mathbf{X}_r^{\mathrm{T}} \mathbf{V}_{r} - \mathbf I_M) \\
& + \gamma(\mathbf{V}_{t}^{\mathrm{T}} \mathbf{X}_t \mathbf{W}_{t}^{\mathrm{T}}\mathbf{W}_{t} \mathbf{X}_t^{\mathrm{T}} \mathbf{V}_{t} - \mathbf I_M)
\end{aligned}
\label{Eq5}
\end{equation}
where $\lambda$ and $\gamma$ are the Lagrange coefficients.

Finding the minimum of $\Pi$ equals to obtain the optimal $\mathbf W_r$, $\mathbf W_t$, $\mathbf V_r$, and $\mathbf V_t$. As a closed-form solution cannot be directly calculated, a two-step iterative optimization method has been detailed in Algorithm 1, which conducts DTW to obtain initial $\mathbf W_r$ and $\mathbf W_t$, and then calculates $\mathbf V_r$ and $\mathbf V_t$ according to Eq. (\ref{Eq5}). The final results are outputted when the value of $\Pi$ converges to a defined value, e.g., 0.01 here.

\subsubsection{Cycling Synchronization for Training Cycles}
Assuming that a source battery is cycled to the end of life, the varying-time trajectories of a source battery consisting of $C$ cycles are denoted as $\mathbf X_1 \in \mathbb{R}^{K_1 \times J}, \mathbf X_2 \in \mathbb{R}^{K_2 \times J}, \cdots, \mathbf X_C \in \mathbb{R}^{K_C \times J}$. The first cycling data $\mathbf X_1$ typically has the longest length, and it is treated as the reference data $\mathbf X_r$ of Eq. (\ref{Eq5}). EDTW allows temporal synchronization between sequential cycles with $\mathbf X_r$. According to Algorithm 1, the synchronized trajectory of $\tilde {\mathbf X}_c$ will be arranged as below:
\begin{equation}
\tilde {\mathbf X}_c = \mathbf V_{t,c}^{\mathrm{T}} \mathbf X_c  \mathbf W_{t,c}^{\mathrm{T}} \space \quad c \in [1, C]
\label{Eq6}
\end{equation}
where $\mathbf V_{t,c}^{\mathrm{T}}$ and $\mathbf W_{t,c}^{\mathrm{T}}$ stand for the calculated spatial subspace and temporal subspace for the $c^{th}$ cycle, respectively.

Through the derived spatial and temporal subspaces, each raw cycling data with the variable structure are synchronized into a carefully designed structure. In this new structure, temporal correlation, spatial correlation, and energy discrepancy of the raw data are retained to avoid information loss.

\subsection{Time-Attention SOH Estimation Model}\label{Section3.2}
With the assistance of cycling synchronization, this section introduces the time attention mechanism to distinguish the important sampling times from the less important sampling times. The basic idea is that samplings do not equally contribute to the final estimation. The estimation is more likely to be determined by the sampling times with large variations over cycles, responding to the discrepancy over cycles. In contrast, the inclusion of unimportant samples may weaken the estimation accuracy.

\begin{figure}[!htb]
\centering
\includegraphics[scale=0.4]{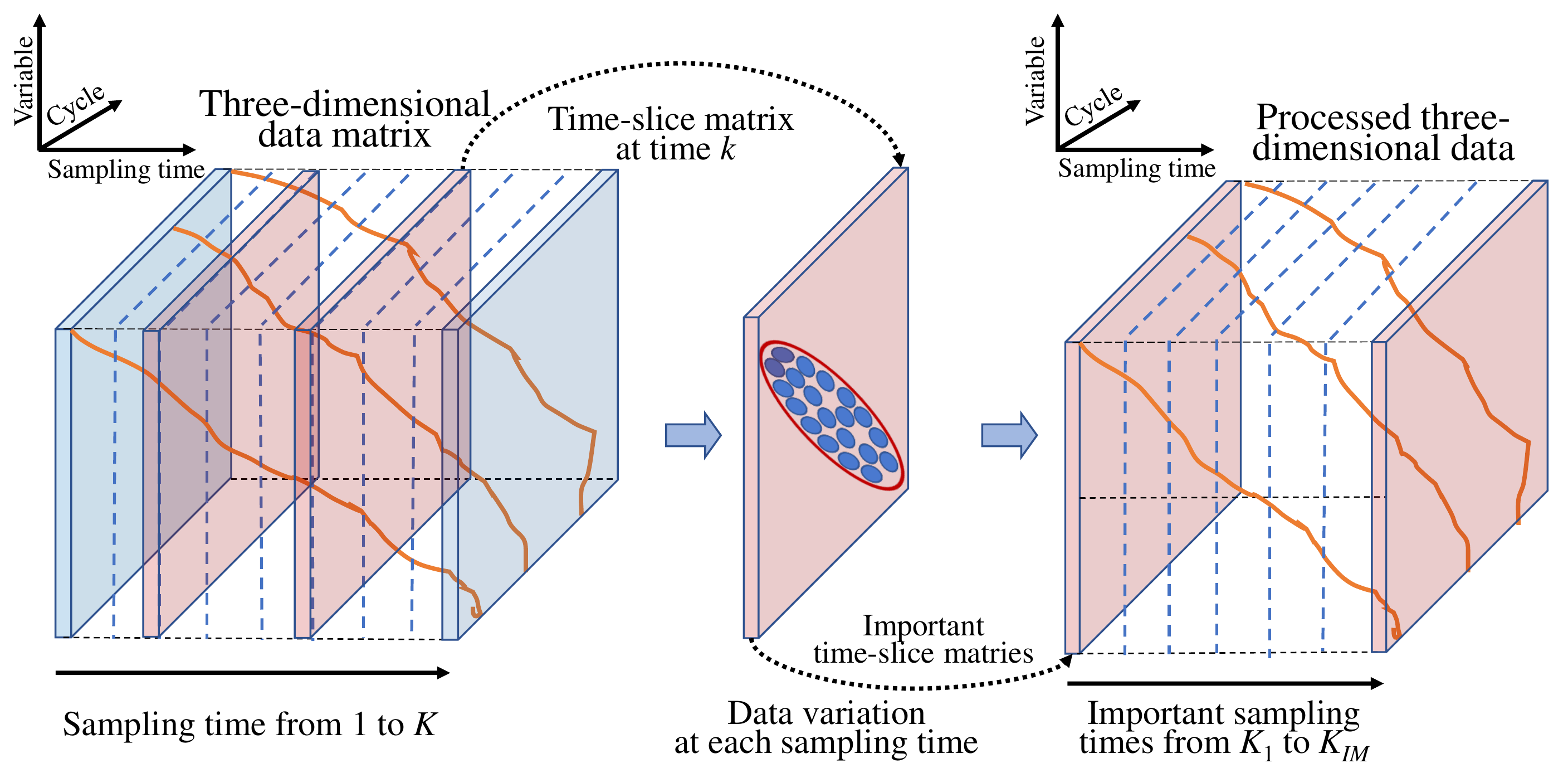}
\caption{Illustration of selecting important sampling times by leveraging time-slice data matrices.}
\label{FIG5}

\end{figure}
\begin{figure}[!htb]
\centering
\includegraphics[scale=0.8]{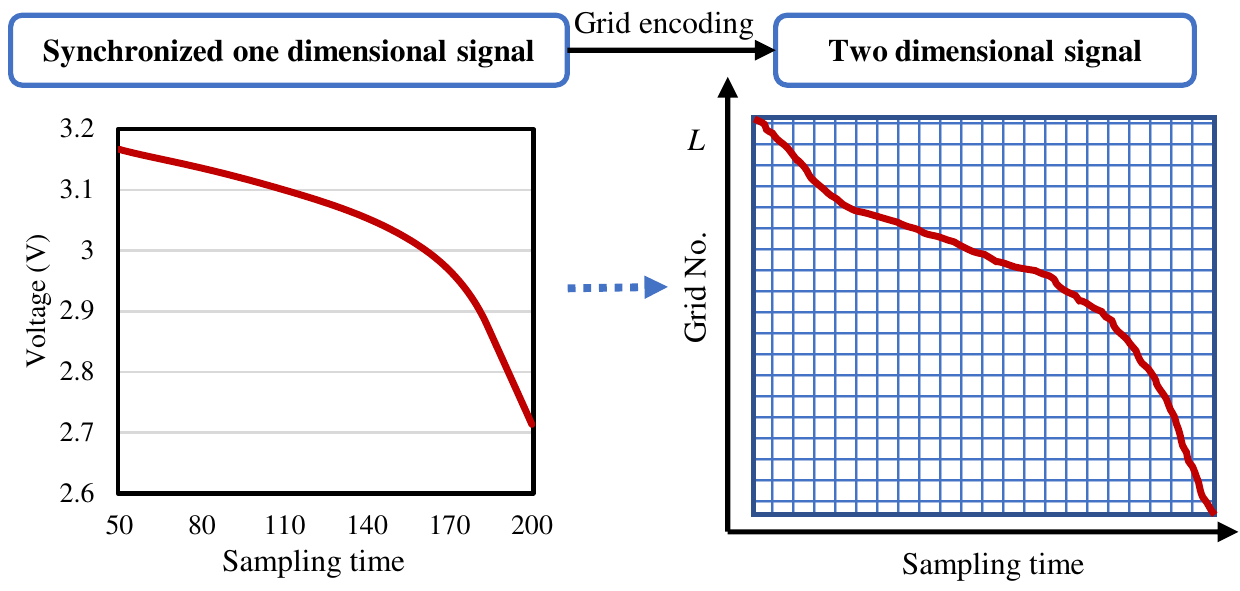}
\caption{Grid representation of time series aiming to magnify the minor difference over cycles.}
\label{FIG6}
\end{figure}

\subsubsection{Determination of Important Sampling Times}
With the consistent cycling data structure, the synchronized trajectories are arranged into a three-dimensional structure to identify the important sampling times. A series of two-dimensional data matrices $\mathbf X_s(k) \in \mathbb{R}^{C \times J}$ are sliced along the time dimension, each of which indicates the data distribution of different trajectories at the same synchronized time interval $k$. Here, we employ the covariance matrix $\mathbf \Lambda(k) = \mathbf X_s^{\mathrm{T}}(k) \mathbf X_s(k)$ to infer the data similarity at time $k$. Noting the maximum covariance is $\Lambda_{max}$, which is calculated by summing the diagonal value of $\Lambda(k)~k\in[1,K]$, the importance degree at each sampling time can be computed as below:
\begin{equation}
I(k) = \frac{\Lambda_k}{\Lambda_{max}} \quad k \in [1, K]
\label{Eq7}
\end{equation}
where $I(k)$ indicates the importance degree at time $k$, and $\Lambda_k$ is the sum of the diagonal value of $\mathbf \Lambda(k)$.

A lower value of $I(k)$ indicates the high similarity of samples over cycles, corresponding to the lower time importance. In contrast, sampling times with significant covariance mean larger time importance.  To distinguish important times from unimportant time steps, it is recommended to use the threshold $\delta$, which is determined via the value of $I(k)$ corresponding to the first knee point of the synchronized discharge voltage. All sampling times are assigned into two groups by comparing with $\delta$ for each trajectory, including $\mathbf X_c^{ID}$ for important samples and unimportant samples, respectively. Furthermore, a three-dimensional data matrix $\mathbf X^{ID} = [\mathbf X_1^{ID}, \mathbf X_2^{ID}, \dots, \mathbf X_C^{ID}]$ is constructed to retain important samples, as shown in Fig. \ref{FIG5}.

\subsubsection{Time-Attention SOH Estimation Model}
Spatial encoding is facilitated on $\mathbf X^{ID}$ to enlarge the minor discrepancy over cycles, especially for the early cycling data. Assuming the scale of an original variable is $[V_{min}, V_{max}]$, the scope of Grid $l \in [1, L]$ is defined as $[V_{min} + (l-1)* \frac{V_{max}-V_{min}}{L}, V_{min} + l* \frac{V_{max}-V_{min}}{L}]$, where $L$ is the number of scaled grid. For cycling data $\mathbf X_c \in \mathbb{R}^{K_{ID} \times J}$, the variable at time $k$ is assigned into one of the grids, and its value is encoded to be 1, as shown in Fig. \ref{FIG6}. For brevity, the encoded cycling data matrix is still denoted as $\mathbf X_c^{ID} \in \mathbb{R}^{JL \times K_{ID}}$.

\begin{figure}[!htb]
\centering
\includegraphics[scale=0.6]{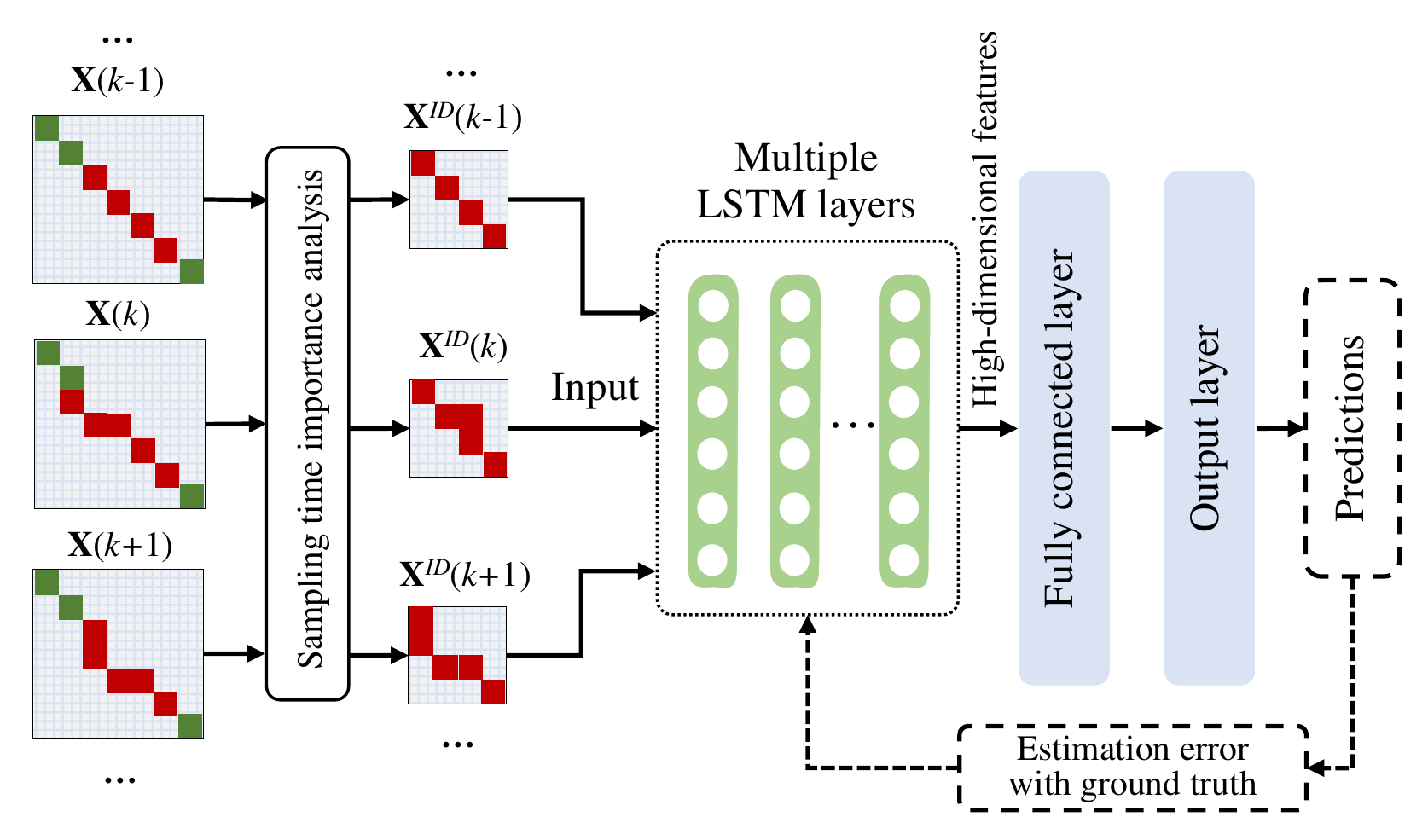}
\caption{Network structure of the time-attention SOH estimation model.}
\label{FIG7}
\end{figure}

The LSTM-based temporal estimation model is adopted to relate $\mathbf X^{ID}$ and actual capacity $\mathbf H$, as shown in Fig. \ref{FIG7}. The selection of LSTM as the backbone network attributes to its superior ability in temporal learning. As LSTM has been widely studied and applied in the topic of SOH estimation, the detailed information regarding LSTM is not repeated here, but can be easily found in \cite{LSTMGAN}-\cite{LSTMMulReed}.

$\mathbf X_c^{ID}$ is fed into a multi-layer LSTM network to regress on the labels $\mathbf H$. The network parameters are optimized by minimizing the estimation errors evaluated by the root mean squared error ($RMSE$) as follows:
\begin{equation}
min E_{trn} = \sqrt{\frac{1}{I} \sum_{i=c}^{I} (\hat{H}_{c} - H_{c})^2}
\label{Eq8}
\end{equation}
where
\begin{equation}
\hat{H}_{c} = F_{LSTM} (\mathbf x_{c}^{ID}; \Theta_l)
\label{Eq9}
\end{equation}
and $\hat{H}_c$ denotes the estimated value of $c^{th}$ cycle, ${H}_{c}$ is the corresponding ground truth, $F_{LSTM}(\cdot)$ stands for the target LSTM model, and $\Theta_l$ is the collection of the parameters used in $F_{LSTM}(\cdot)$.

\subsection{Future Data Reconstruction for Real-time SOH Estimation}\label{Section3.3}
A reliable real-time SOH estimation for online trajectory $\mathbf X_{T}(1:k)$, which are the cycling data from the beginning up to the time step $k$ in a new cycle, requires the estimation of an unknown future trajectory. Here, we attempt to supplement the unknown future trajectory through similarity checking. Afterward, the real-time SOH estimation is performed through two sequential components: online cycling synchronization and real-time estimation.

\subsubsection{Data Supplementation and Online Cycling Synchronization}
Similarity evaluation is performed between $\mathbf X_{T}(1:k)$ and each cycling trajectory from the training dataset up to the same time step. The similarity is calculated using the Euclidean distance between $\mathbf X_{T}(1:k)$ and the candidate trajectories, which is shown as follows:
\begin{equation}
S_c(k) = \sqrt{\sum_{k=1}^{K} (\mathbf X_{T}(1:k) - \mathbf X_{c}(1:k))^2}
\label{Eq10}
\end{equation}
where $\mathbf X_{c}(1:k)$ and $\mathbf X_{T}(1:k)$ are the samples from $1$ to $k$ of original cycling data $\mathbf X_{c}$ and $\mathbf X_{T}(k)$, respectively. It is worth noting that if the value of $k$ is larger than the length of partial trajectories from training data, these trajectories are naturally out of consideration. Sequentially, the specifics for online cycling synchronization are given as follows:

\begin{itemize}
\item Step 1: For the given $\mathbf X_T(1:k)$, a series of similarity $\mathbf S_k=[S_1(k), S_2(k), \dots, S_C(k)]$ are derived with Eq. (\ref{Eq10}) for all training cycles. The matched future trajectory is selected from the cycle with the minimum of $\mathbf S_k$;
\item Step 2: Assuming that the selected trajectory is $\mathbf X_s$, the reconstructed data are combination of $\mathbf X_{T}(1:k)$ and $\hat {\mathbf X}_{s}(k+1:K_s)$, e.g., $\hat {\mathbf X}_{T} = [\mathbf X_{T}(1:k), \hat {\mathbf X}_{s}(k+1:K_s)]$;
\item Step 3: Performing temporal synchronization on $\hat {\mathbf X}_{T}$ and the synchronized data $\tilde {\mathbf X}_{T}$ will be computed with  Eq. (\ref{Eq6}) according to {Algorithm 1};
\end{itemize}

\subsubsection{Real-time SOH Estimation}
The online SOH estimation can be calculated with the following steps:
\begin{itemize}
\item Step 4: Retaining important samples of $\tilde {\mathbf X}_{T}$ and the new matrix is denoted as $\tilde {\mathbf X}_{T}^{ID}$;
\item Step 5: Conducting spatial encoding on $\tilde {\mathbf X}_{T}^{ID}$;
\item Step 6: Obtaining the real-time estimation at time $k$ with the well-trained model given in Section II.B as follows:
\begin{equation}
\hat{H}_{T}(k) = \Psi_{LSTM} (\tilde {\mathbf X}_{T}^{ID}; \mathbf \Theta)
\label{Eq11}
\end{equation}
\end{itemize}

Updating ${\mathbf X}_{T}(1:k)$ to ${\mathbf X}_{T}(1:k+1)$ with the time increase and processing the updated data  ${\mathbf X}_{T}$ according to Steps 1 through 6 yields continuous SOH estimation on the fly. The iteration procedure stops when the end of a discharging cycle is reached. The estimation at the initial stage of a cycle will be inaccurate, as the future data account for a large percentage. Naturally, the estimation will become increasingly reliable and close to the ground truth with time increase, especially at the important sampling times.

\textit{Remark:} It is worth noting that model updating is a matter when the well-trained offline model has been used for dissimilar batteries, or new working conditions \cite{TransLearing}. Transfer learning is recommended to handle such an issue when the data distribution is distinct between the training and testing domains. As such, paying attention to this issue is promising in future work.

\section{Experimental Results and Discussion}\label{Section4}
This section provides extensive comparisons to demonstrate the advantages of variable cycling data synchronization, time-attention modeling, and real-time SOH estimation, verifying the efficacy of the proposed method with multiple battery cells.

\begin{table}[!ht]
\renewcommand{\arraystretch}{1.5}
\centering
\caption{Grouping of all battery cells according to the number of cycle length.}
\scriptsize
\begin{tabular}{p{1.8cm} p{3.8cm} p{1.8cm}}
\hline
\hline
\textbf{Cycle length} & \textbf{Grouped batteries} & \textbf{Test battery} \\
\hline
{$\leq$ 600} & {17, 19, 33, 34, 35, 39, 40, 43}  & 17 \\
\hline
{601 to 700} & {14, 16, 18, 23, 24, 30, 38} & 14 \\
\hline
{701 to 800} & {2, 3, 7, 8, 11, 12, 13, 22, 29, 47} & 2 \\
\hline
{801 to 900} & {1, 10, 20, 21, 26, 36, 37, 41, 44, 6} & 1 \\
\hline
{901 to 1000} & {9, 15, 25, 27, 28, 32} & 9 \\
\hline
{$\geq$ 1001} & {31, 42, 45, 48} & 31 \\
\hline
\hline
\end{tabular}
\label{TABLE1}
\end{table}

\begin{table}[!ht]
\renewcommand{\arraystretch}{1.5}
\centering
\caption{Comparison between the proposed method and its counterparts by adopting Battery 22 as the reference data.}
\scriptsize
\begin{center}
\begin{threeparttable}
\begin{tabular}{p{3cm} p{1.5cm} p{1cm} p{1cm} p{1cm} p{1cm} p{1cm} p{1cm} p{1cm} p{2cm}}
\hline
\hline
\multirow{2}{*}{\textbf{Method}} & \multirow{2}{*}{\textbf{Index}} & \multicolumn{7}{c}{\textbf{Battery No.}} & \multirow{2}{*}{\textbf{Average performance}} \\
\cline{3-9}
& & 1 & 2 & 9 & 14 & 16 & 17 & 31 & \\
\hline
\multirow{2}{*}{The proposed method} & $RMSE (\%)$ & \textbf{0.97} & \textbf{0.74} & \textbf{1.08} & \textbf{1.02} & \textbf{1.66} & \textbf{1.43} & \textbf{0.64} & \textbf{1.08 $\pm$ 0.36} \\
& $R^2 (\%)$ & \textbf{98.78} & \textbf{99.41} & \textbf{98.96} & \textbf{98.86} & \textbf{95.03} & \textbf{96.40} & \textbf{98.80} & \textbf{98.03 $\pm$ 1.65} \\
\hline
\multirow{2}{*}{LSTM1 (Manual trunction)} & $RMSE (\%)$ & 3.99 & 3.74 & 4.03 & 3.58 & 2.69 & 5.08 & 2.95 & 3.72 $\pm$ 0.76 \\
& $R^2 (\%)$ & 79.08 & 84.82 & 85.44 & 86.05 & 86.68 & 50.67 & 73.84 & 78.05 $\pm$ 12.93 \\
\hline
\multirow{2}{*}{LSTM2 (Linear interpolation)} & $RMSE (\%)$ & 1.43 & 1.62 & 5.10 & 2.02 & 2.66 & 3.26 & 2.33 & 2.63 $\pm$ 1.25 \\
& $R^2 (\%)$ & 87.16 & 97.14 & 76.69 & 95.50 & 87.02 & 79.69 & 83.66 & 86.70 $\pm$ 7.59 \\
\hline
\multirow{2}{*}{SVM} & $RMSE (\%)$ & 2.34 & 2.53 & 3.38 & 2.51 & 1.83 & 2.23 & 1.99 & 2.40 $\pm$ 0.50 \\
& $R^2 (\%)$ & 92.79 & 93.05 & 89.72 & 93.03 & 93.86 & 90.49 & 88.07 & 91.57 $\pm$ 2.16 \\
\hline
\hline
\end{tabular}
\end{threeparttable}
\end{center}
\label{TABLE2}
\begin{tablenotes}
\item[1] [1] The bold text indicates the best results.
\end{tablenotes}
\end{table}

\begin{figure}[!ht]
	\centering
	\subfigure[]
	{
	\begin{minipage}[t]{0.45\linewidth}
	\centering
	\includegraphics[width=7.5cm]{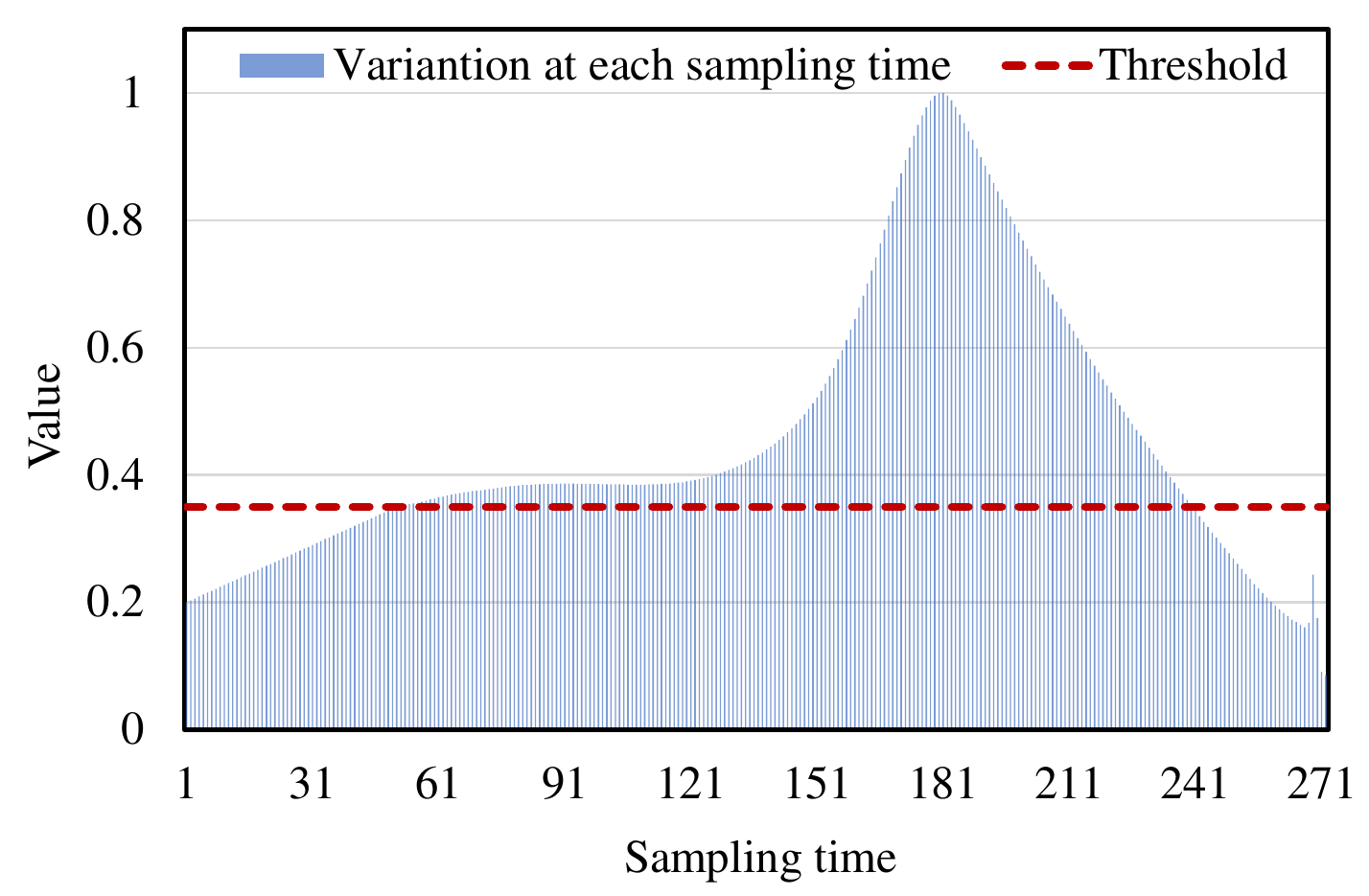}
	\end{minipage}
	}
	\subfigure[]
	{
	\begin{minipage}[t]{0.45\linewidth}
	\centering
	\includegraphics[width=7.5cm]{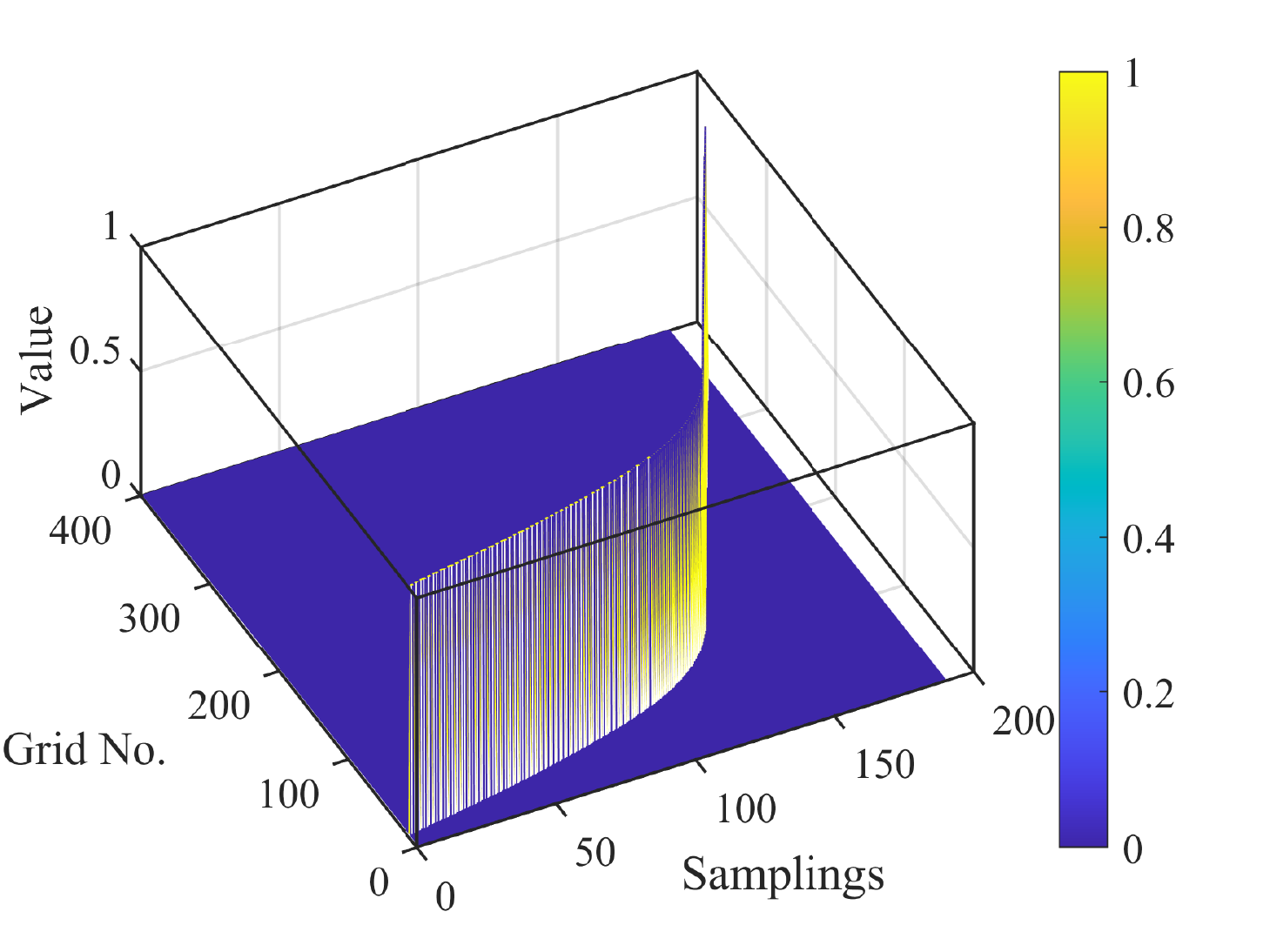}
	\end{minipage}
	}
	\caption{(a) Illustration of important sampling times when the threshold is 0.35, and (b) Visualization of the encoded cycling data of the $301^{th}$ cycle in Battery 22.}
	\label{FIG8}
\end{figure}

In the employed Massachusetts Institute of Technology dataset \cite{RefNature}, each LiFePO4 battery cell is cycled from fresh to retired status using temperature chamber, Arbin test system, and working station. The nominal capacity of deployed cells is around 1.1 Ah, which slightly differs from each other due to subtle manufacturing inconsistency. During cycling tests, a cell repetitively experiences fast-charging/discharging. The state-of-charge of a battery is charged to 80$\%$ with four sequential constant currents and fully charged to the constant voltage with 3.6V.

\begin{figure}[!ht]
	\centering
	\subfigure[Battery 1]
	{
	\begin{minipage}[t]{0.47\linewidth}
	\centering
	\includegraphics[width=8.6cm]{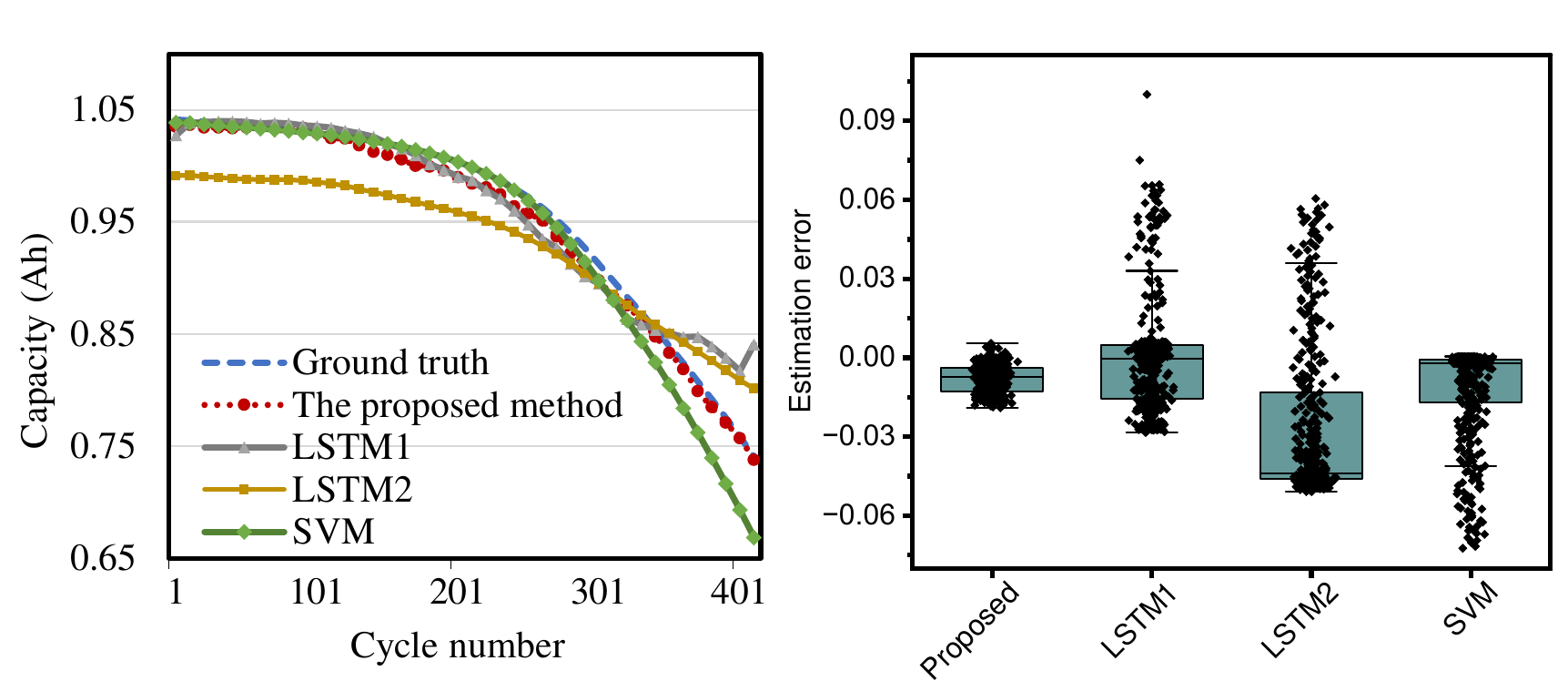}
	\end{minipage}
	}
	\subfigure[Battery 2]
	{
	\begin{minipage}[t]{0.47\linewidth}
	\centering
	\includegraphics[width=8.6cm]{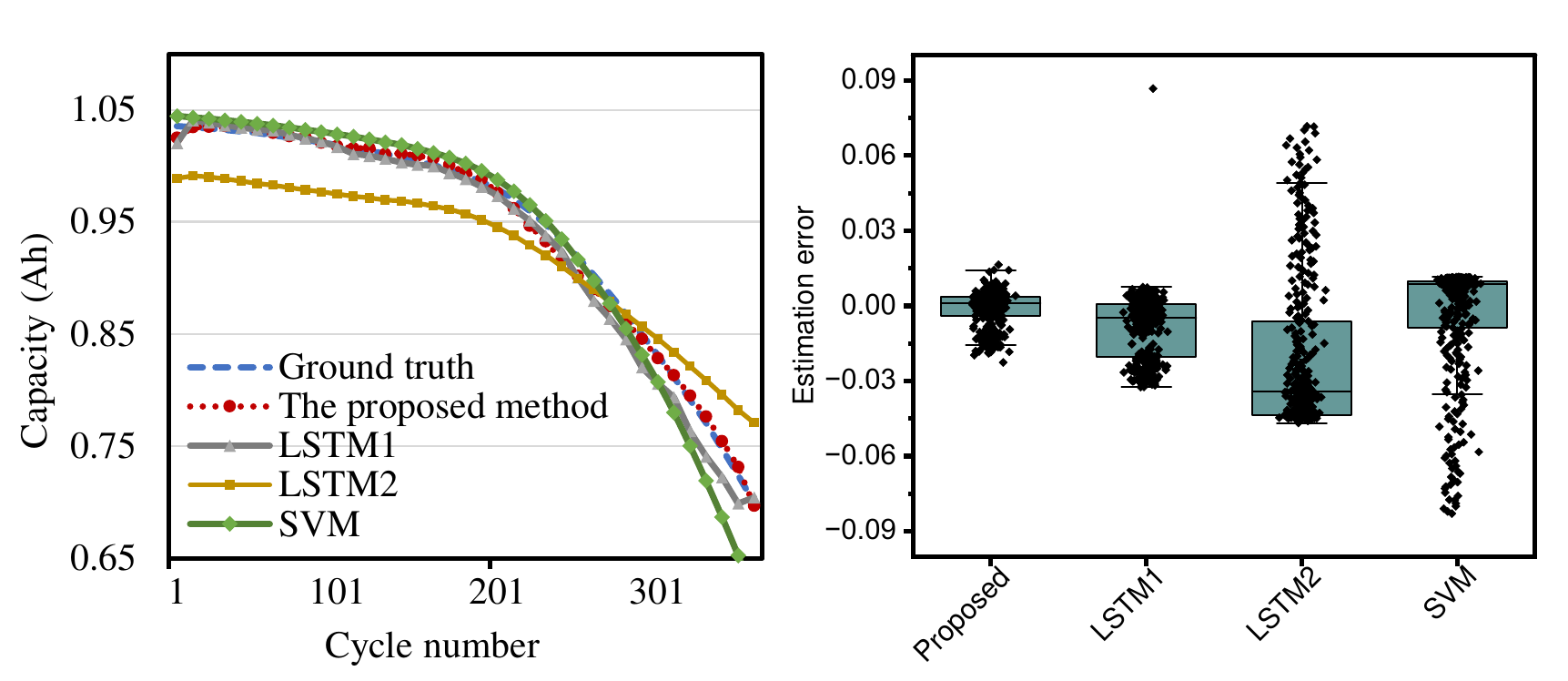}
	\end{minipage}
	}

	\subfigure[Battery 9]
	{
	\begin{minipage}[t]{0.47\linewidth}
	\centering
	\includegraphics[width=8.6cm]{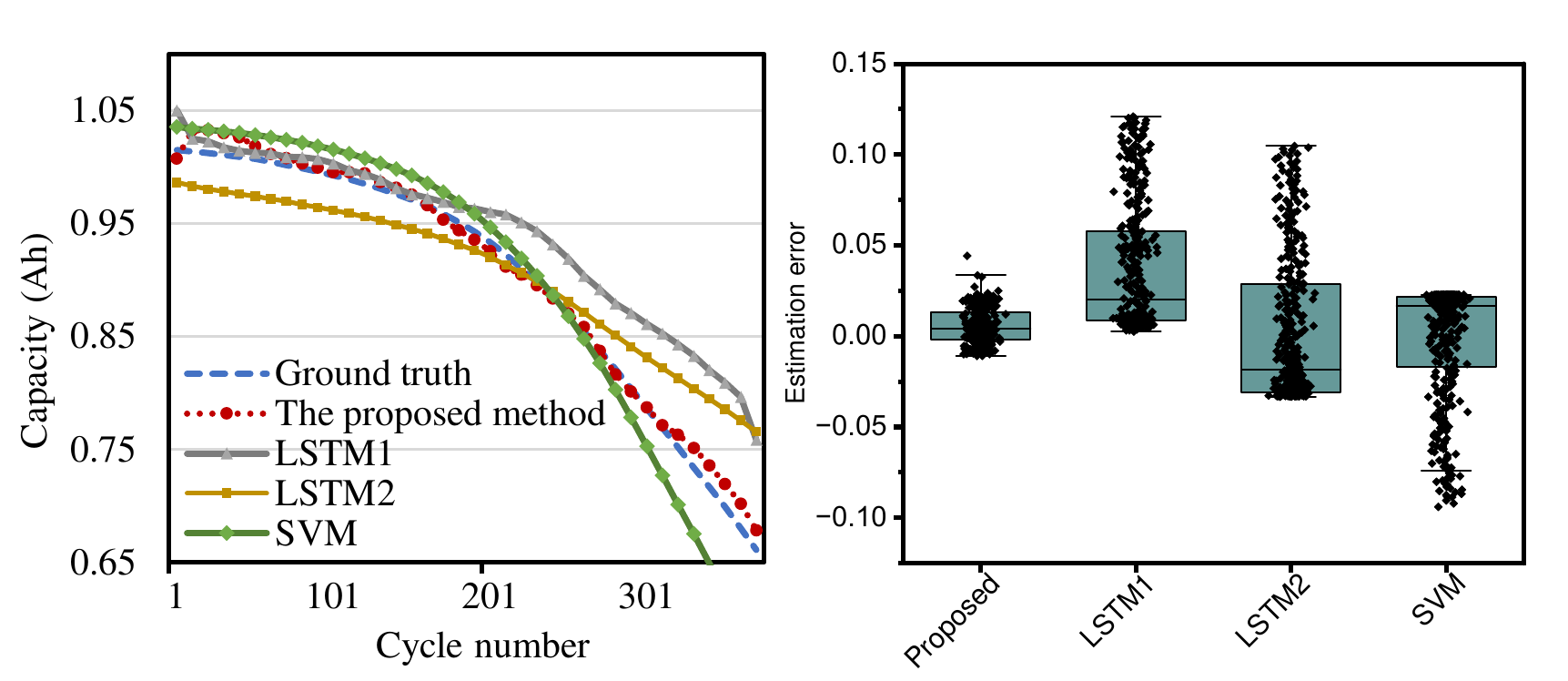}
	\end{minipage}
	}
	\subfigure[Battery 14]
	{
	\begin{minipage}[t]{0.47\linewidth}
	\centering
	\includegraphics[width=8.6cm]{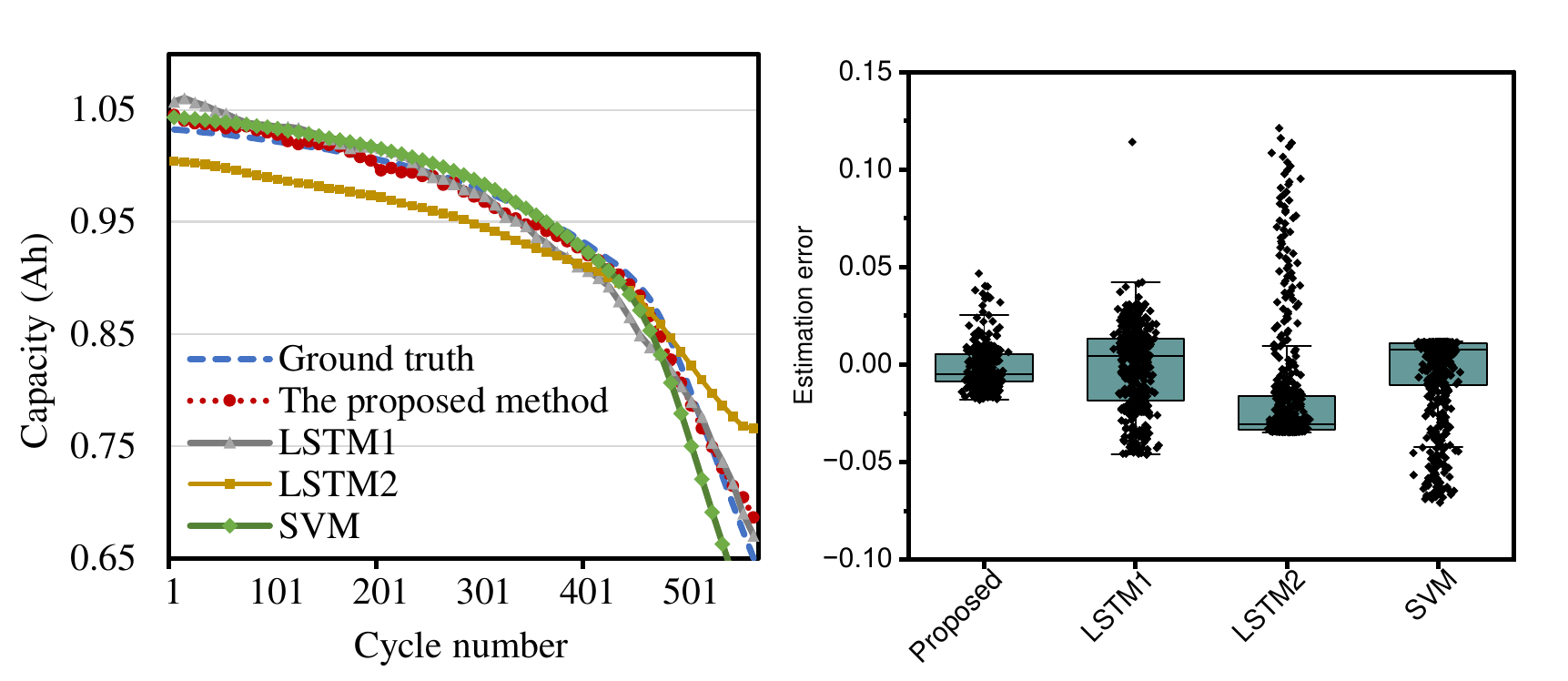}
	\end{minipage}
	}

	\subfigure[Battery 17]
	{
	\begin{minipage}[t]{0.47\linewidth}
	\centering
	\includegraphics[width=8.6cm]{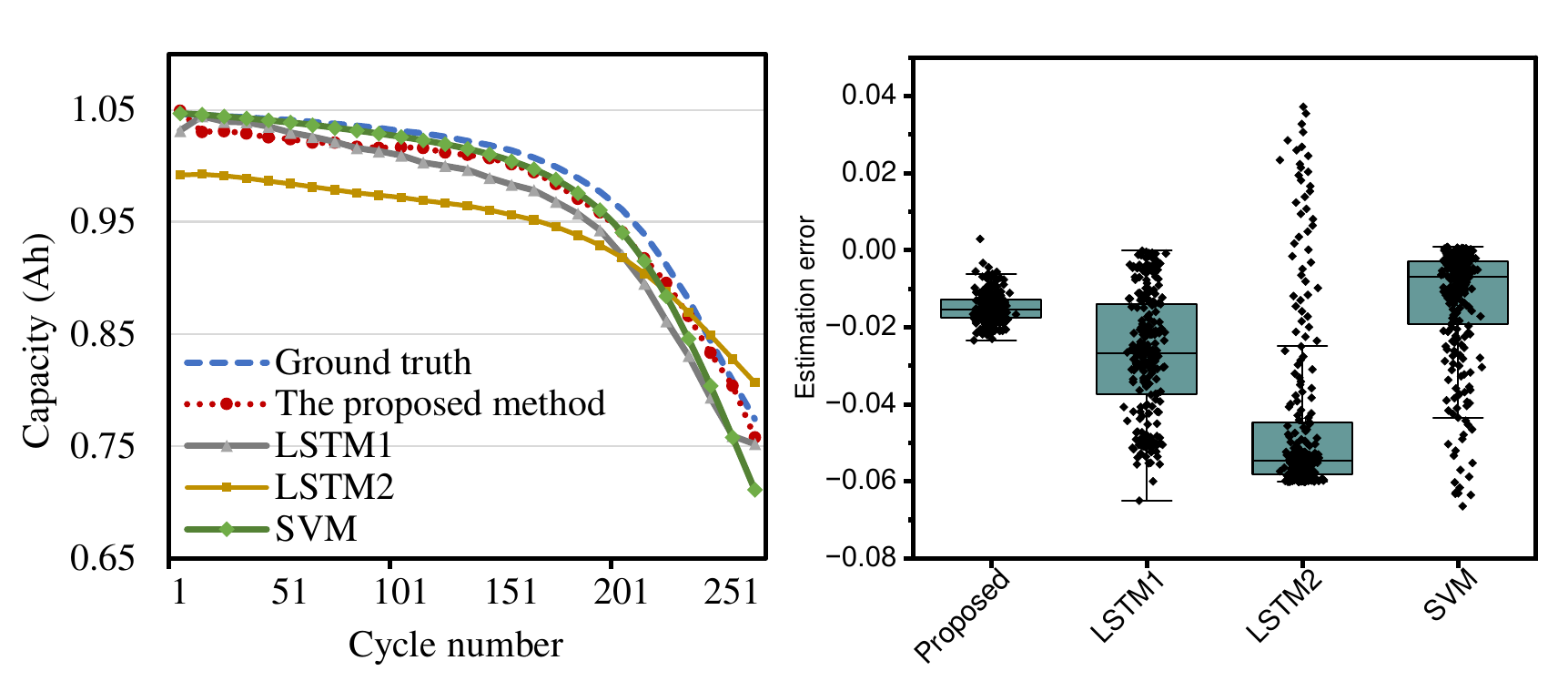}
	\end{minipage}
	}
	\subfigure[Battery 31]
	{
	\begin{minipage}[t]{0.47\linewidth}
	\centering
	\includegraphics[width=8.6cm]{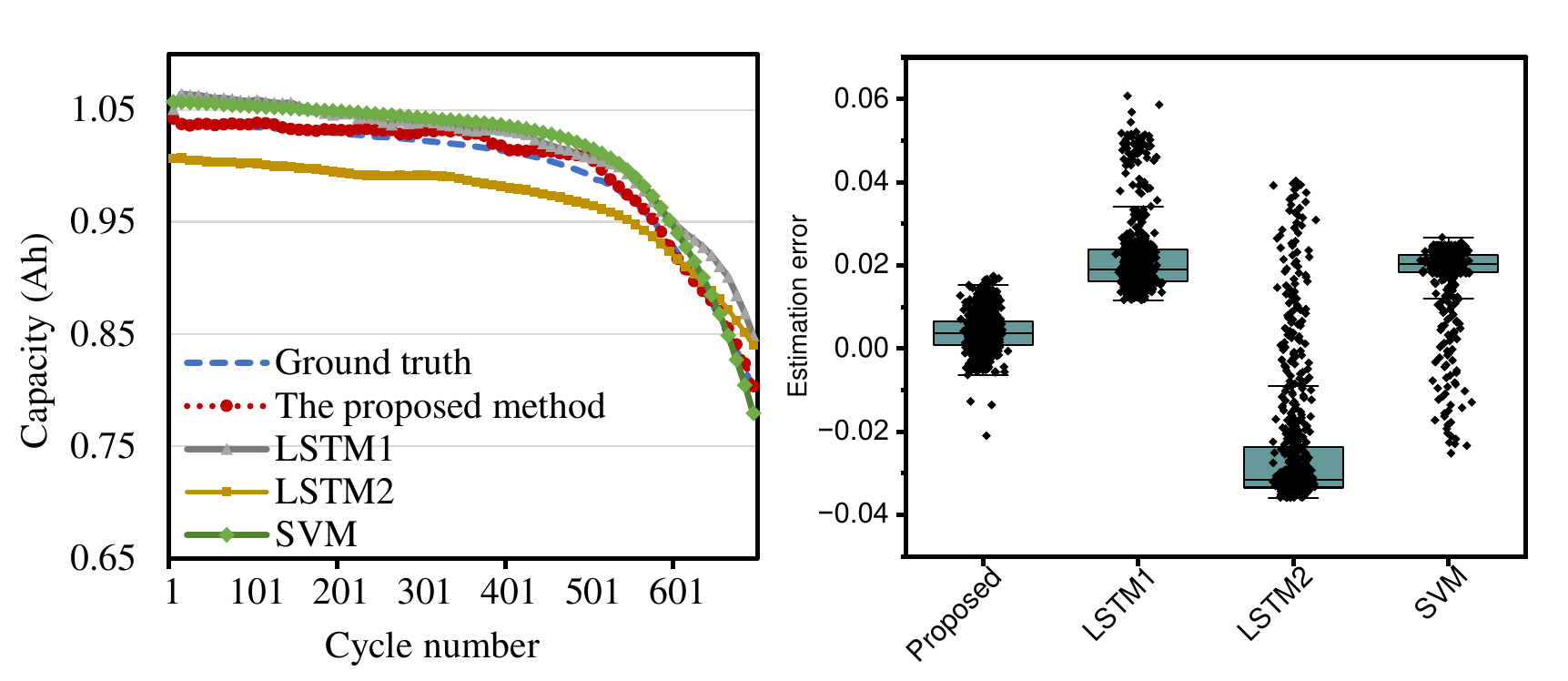}
	\end{minipage}
	}
	\caption{The comparisons between the proposed method and its counterparts regarding estimations and errors.}
	\label{FIG9}
\end{figure}

\begin{figure}[!ht]
	\centering
	\subfigure[The 301 $^{th}$ cycle]
	{
	\begin{minipage}[t]{0.3\linewidth}
	\centering
	\includegraphics[width=6cm]{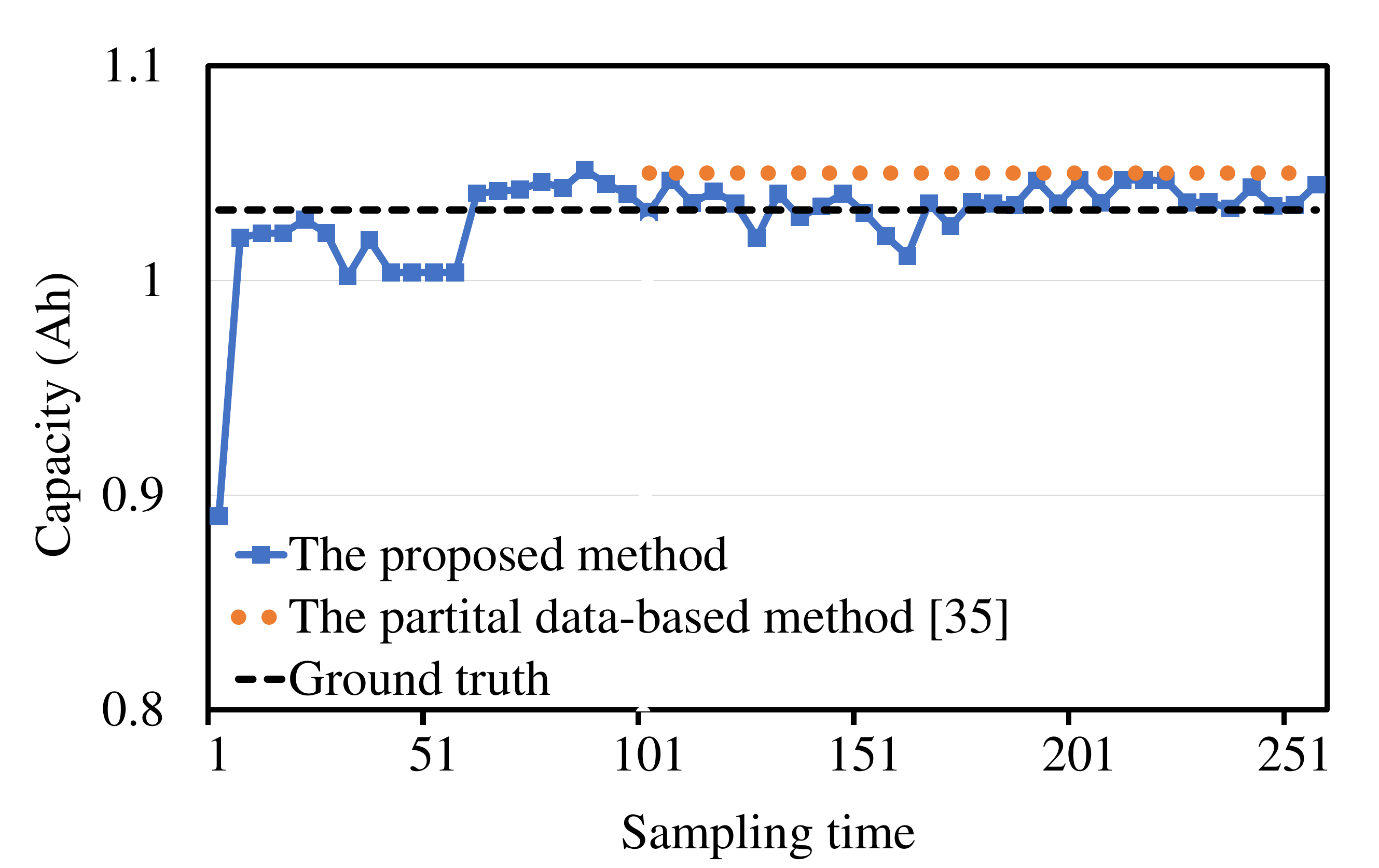}
	\end{minipage}
	}
	\subfigure[The 401 $^{th}$ cycle]
	{
	\begin{minipage}[t]{0.3\linewidth}
	\centering
	\includegraphics[width=6cm]{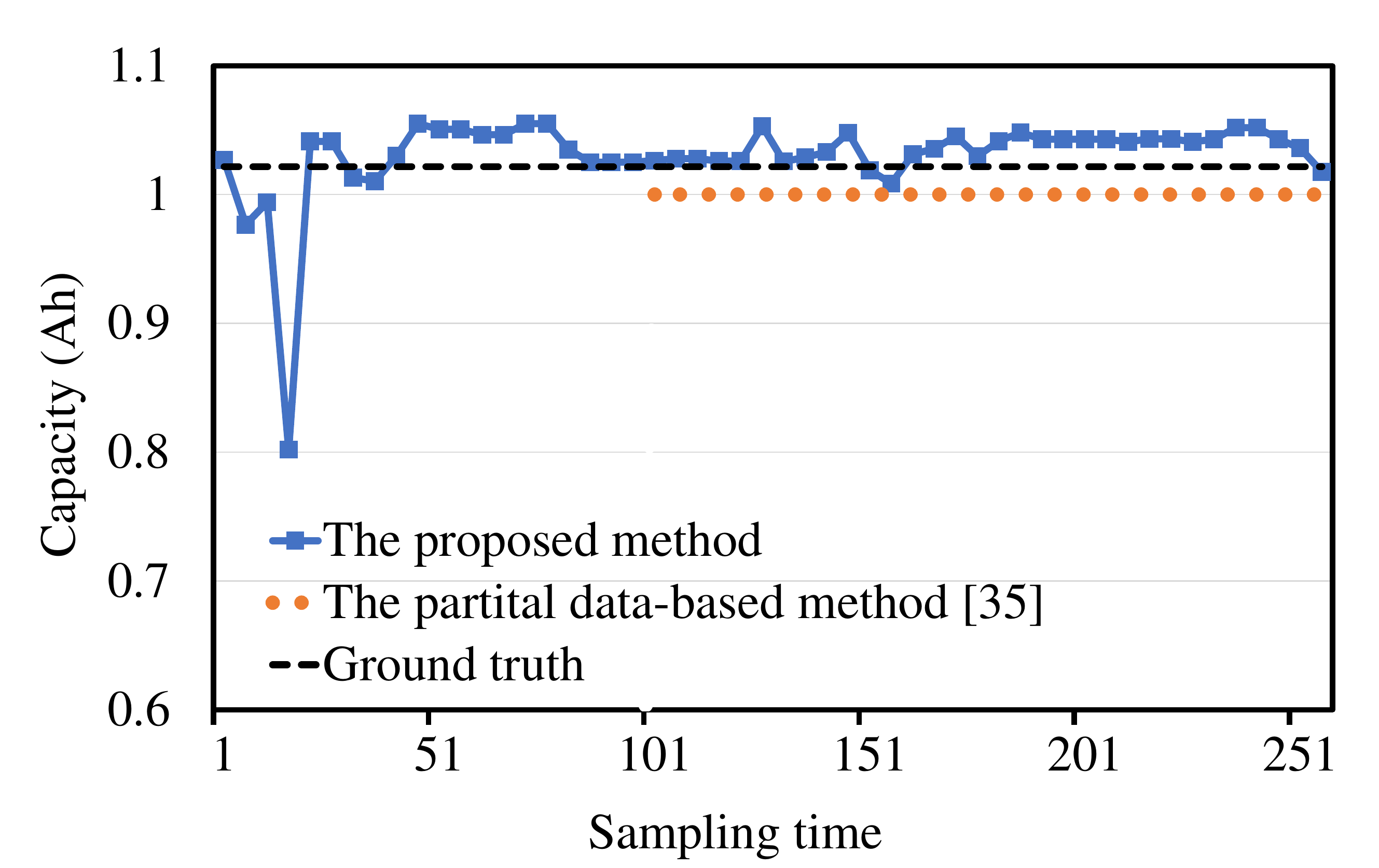}
	\end{minipage}
	}
	\subfigure[The 501 $^{th}$ cycle]
	{
	\begin{minipage}[t]{0.3\linewidth}
	\centering
	\includegraphics[width=6cm]{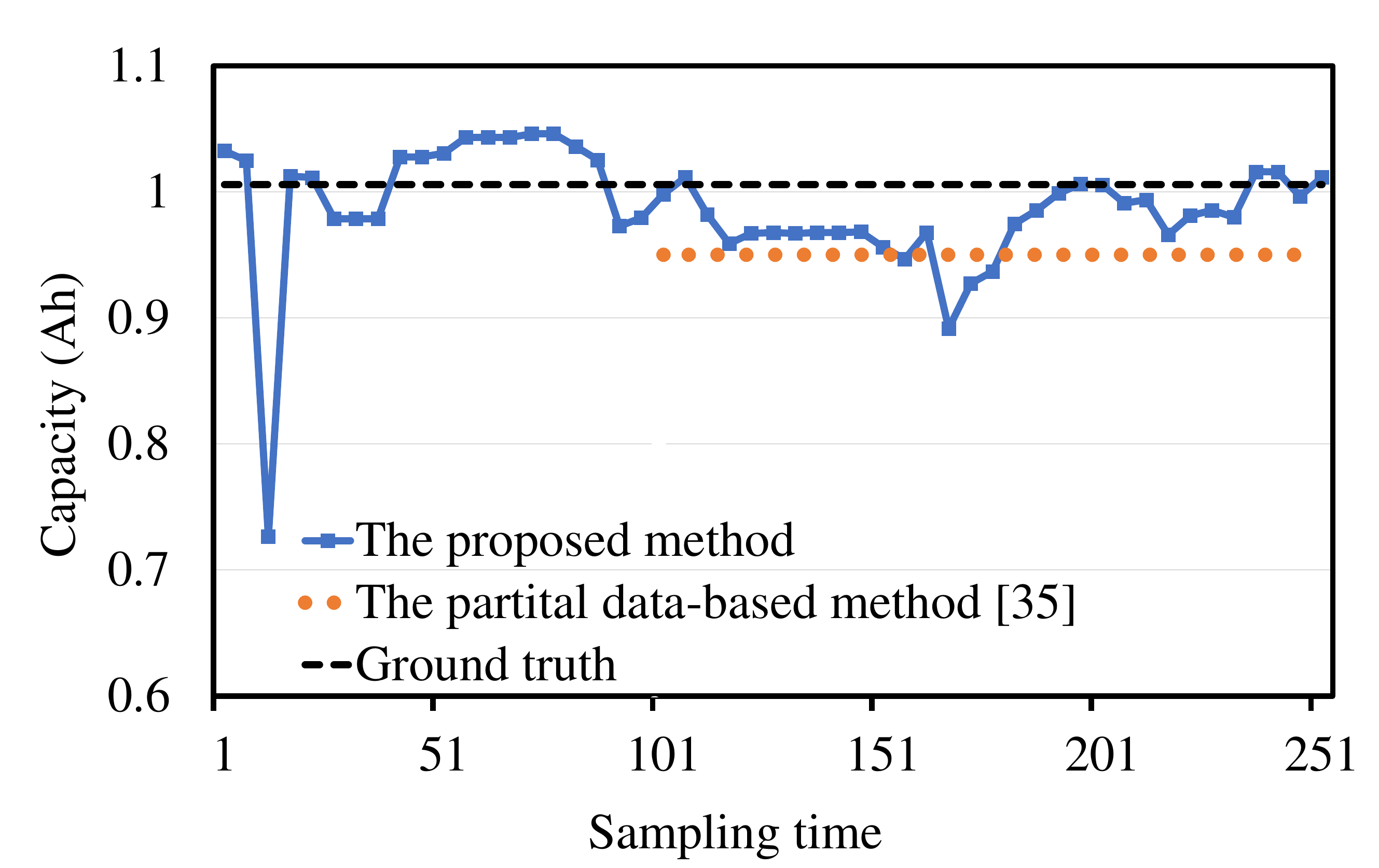}
	\end{minipage}
	}
	
	\subfigure[The 601 $^{th}$ cycle]
	{
	\begin{minipage}[t]{0.3\linewidth}
	\centering
	\includegraphics[width=6cm]{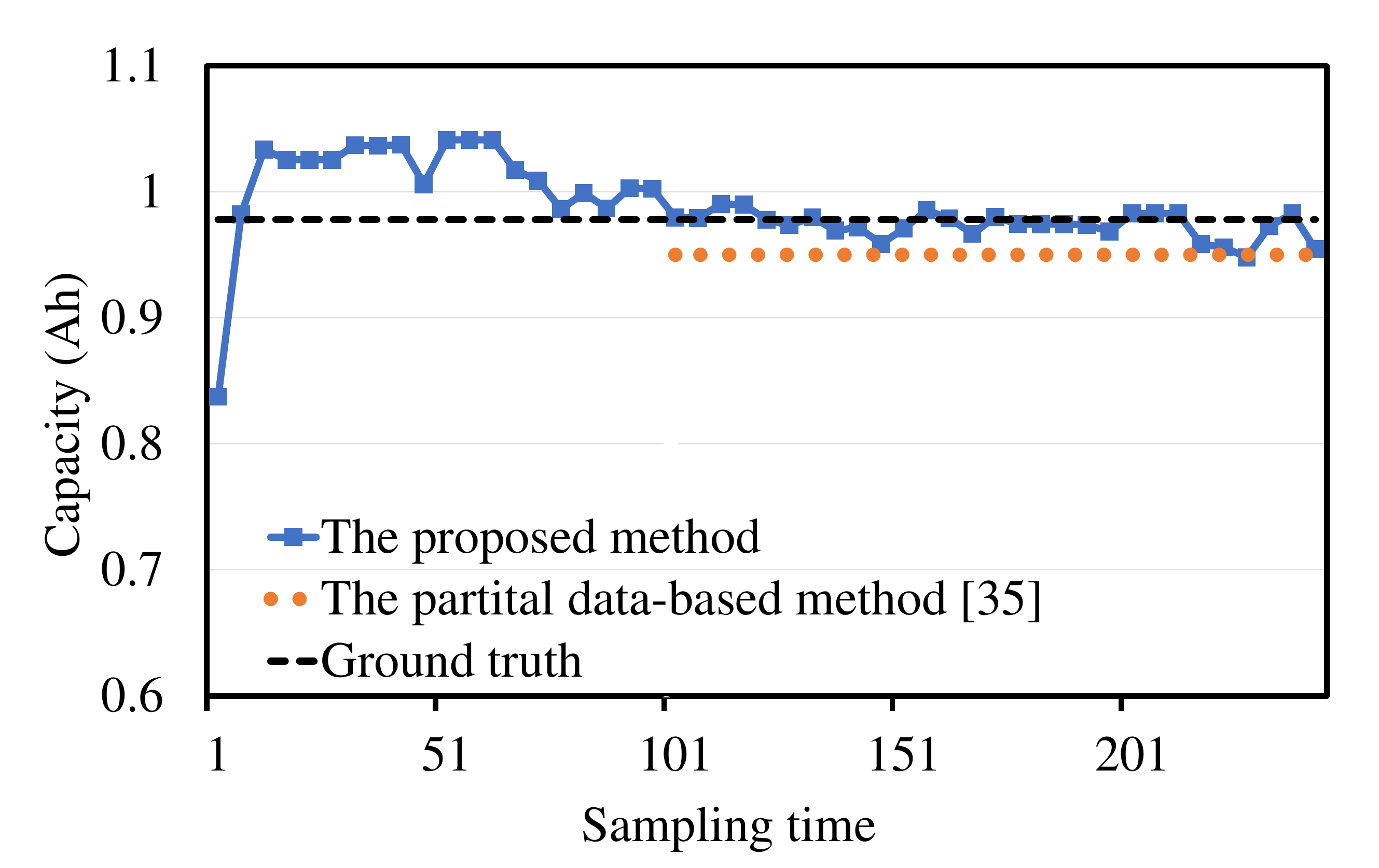}
	\end{minipage}
	}
	\subfigure[The 701 $^{th}$ cycle]
	{
	\begin{minipage}[t]{0.3\linewidth}
	\centering
	\includegraphics[width=6cm]{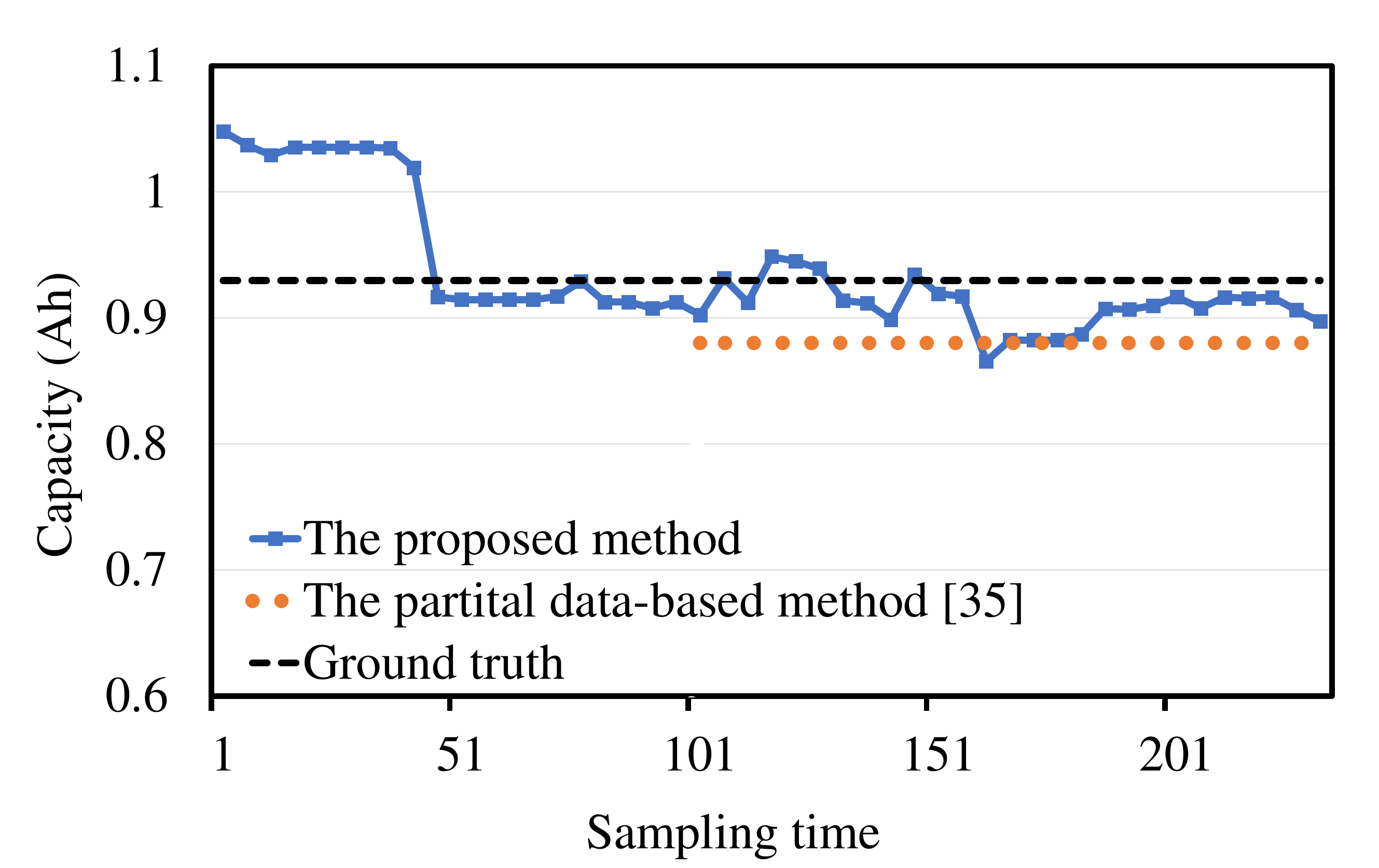}
	\end{minipage}
	}
	\subfigure[The 801 $^{th}$ cycle]
	{
	\begin{minipage}[t]{0.3\linewidth}
	\centering
	\includegraphics[width=6cm]{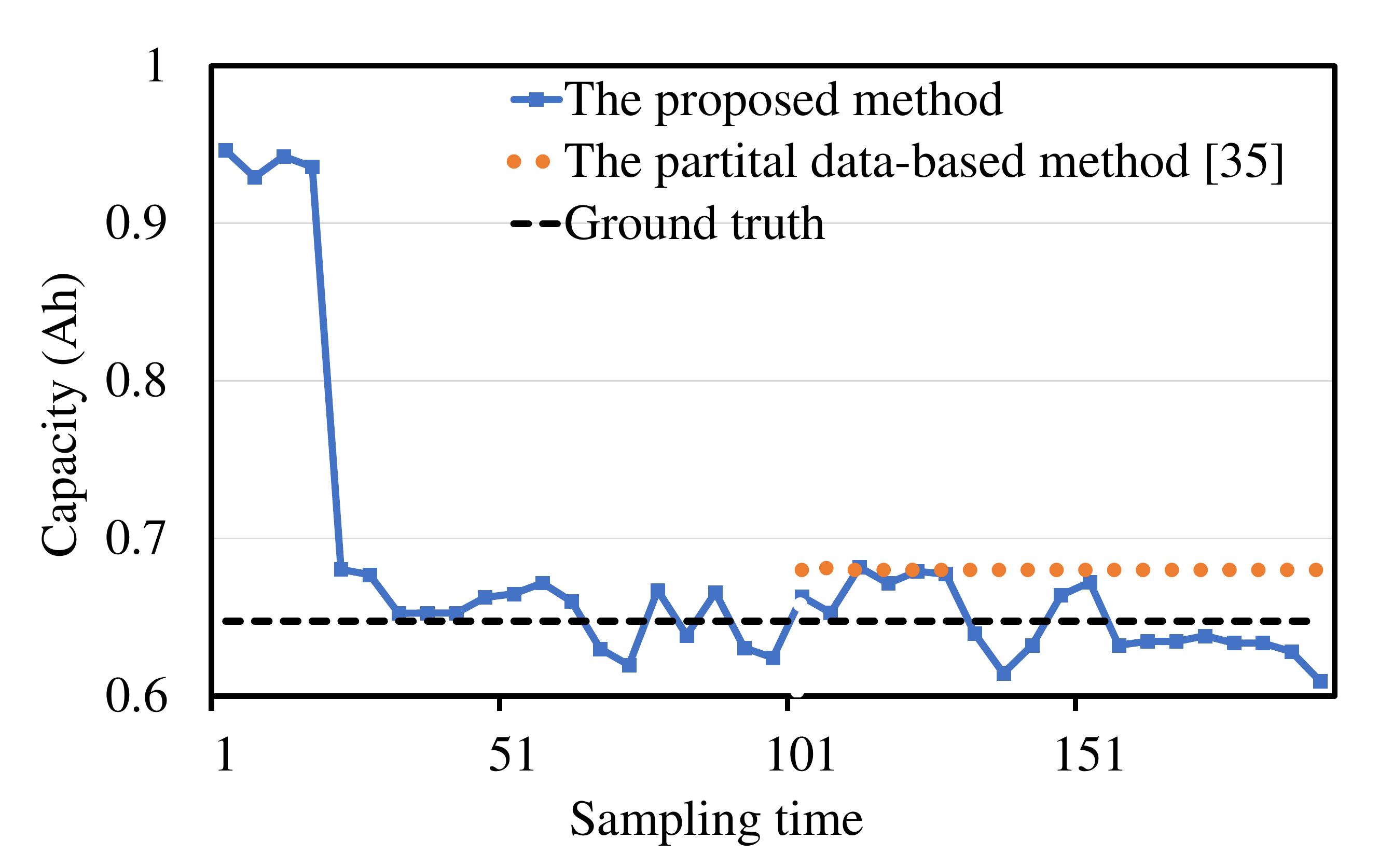}
	\end{minipage}
	}
	\caption{The real-time estimation results regarding Battery 1 using the proposed method with an interval of 100 cycles.}
	\label{FIG10}
\end{figure}

\begin{figure}[!ht]
	\centering
	\begin{minipage}[t]{1\linewidth}
	\centering
	\includegraphics[width=8cm]{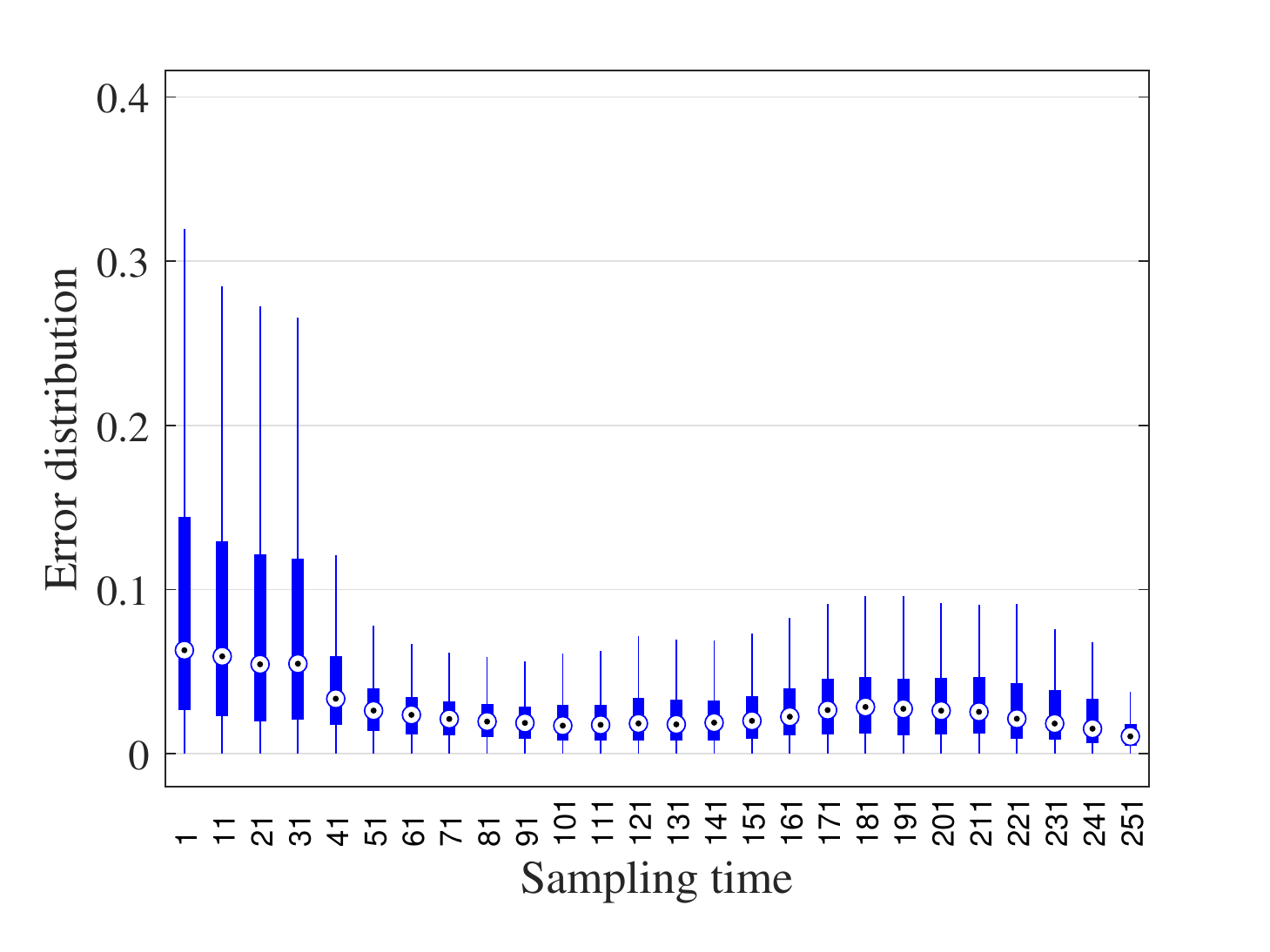}
	\end{minipage}
	\caption{Visualization of estimation error at each sampling time over cycles for Battery 1.}
	\label{FIG11}
\end{figure}

In this study, we use the discharging data to estimate the capacity, including battery temperature and discharging voltage. Through repetitive optimized fast-charging and constant discharging manner, the capacity of each battery will degrade naturally and stop testing around 0.65Ah. This study uses the discharging temperature and voltage to estimate the capacity, but the discharging current is not adopted as it is a constant value. The total number of 45 aging cells is tested for analysis and is manually grouped into six categories according to different cycling lengths. Table \ref{TABLE1} lists detailed information for these LIB cells, and the specific batteries used are highlighted.

\subsection{Variable Cycling Data Synchronization}\label{Section4.1}
Battery 22 is selected as the training battery, as it has a middle lifespan among other batteries with 862 cycles. Here, we take its first discharging cycle as the reference cycle, which facilitates implementation for offline cycling synchronization. And other discharging data are synchronized to the reference one according to steps given in Section \ref{Section3.1}.

\subsection{Time-Attention SOH Estimation Model}\label{Section4.2}
\subsubsection{ Determination of Important Sampling Times}
We calculate covariance at each sampling time with the synchronized data, and obtain a covariance vector with 272 dimensions. Fig. \ref{FIG8}(a) illustrates the important sampling intervals in comparison with a threshold, which is 0.35 calculated from the first knee point. It is observed that samplings from 51 to 240 are identified as essential intervals.

\subsubsection{Time-Attention SOH Estimation Model}
Since the first 300 cycles in Battery 22 donot experience performance degradation, they are regarded as the healthy status and will not be used for modeling. The SOH estimation model is trained with the first 70$\%$ data from cycle 301 to cycle 862, i.e., Cycle 301 to Cycle 652. The remaining 30$\%$ cycles are used for testing.

We develop a time-attention SOH estimation model using a LSTM network with two layers, each of which contains 100 LSTM cells. The first battery from each group listed in Table \ref{TABLE1} is selected for testing the performance of the proposed method, and the well-trained time-attention SOH estimation from Battery 22 is directly applied without network fine-tuning. With EDTW, the cycling data of each testing battery have been offline synchronized with the first cycle of Battery 22 with the length of 272. The important sampling times are from the 51$^{st}$ to 240$^{th}$ intervals in the synchronized cycling data. The selected samples will be further encoded to a two-dimensional matrix with $400 \times 190$ according to the ways given in Section \ref{Section3.3}, as shown in Fig. \ref{FIG8}(b).

Three kinds of conventional SOH estimation models are used for further comparison quantitatively. The first approach is the LSTM-based model with manual truncation, truncating all cycling data more than the shortest cycle length. In contrast, the cycling data are extended to the longest length through linear interpolation for the following two approaches. Specifically, for the second and third models, LSTM and support vector machine (SVM) are employed as the estimation approaches, respectively. Table \ref{TABLE2} summarizes the detailed results for seven battery cells using the indices $RMSE$ and $R^2$ \cite{RealSOH3}. By highlighting the best result in bold format, it is observed that the proposed method outperforms the counterparts for all battery cells largely. In addition, Table \ref{TABLE2} provides the overall performance across all battery cells through the mean and standard variation. Fig. \ref{FIG9} plots the predictions and $RMSE$ error distribution of the first battery from each group, and the proposed method yields the best estimation accuracy for all testing batteries. As such, the efficacy of the proposed method in capacity estimation has been successfully verified in the end-of-cycle estimation.

\subsection{Real-time SOH Estimation Model}
This part further illustrates the real-time SOH estimation performance for Battery 1 at each sampling time over cycles. Future trajectory estimation is conducted to match the most similar trajectory from the reference battery, and thus the mixed data are sequentially synchronized and encoded in the same way as that of the offline estimation. Afterward, the estimation at time $k$ is calculated using the well-trained end-cycle model. From Figs. \ref{FIG10}(a) to \ref{FIG10}(f), a series of results have been demonstrated every 100 cycles from the 300$^{th}$ cycle. For each cycle, the proposed real-time estimation enables sensing the battery's health status on the fly. Moreover, the results are very close to the end-cycle ground truth from the 51$^{st}$ sampling.

Furthermore, the recently published partial data-based SOH method \cite{LiuTVT} is employed for comparison. Using partial data from the complete cycle to infer SOH estimation, \cite{LiuTVT} finds the best starting voltage from a pool of discrete voltages through manual selections. Afterward, the voltage drop during equal time intervals is used as the input to the extreme learning machine. Here, the specified value of starting voltage is determined as 3.20V, and the duration is given as 50 sampling intervals. As such, \cite{LiuTVT} yields the SOH estimation from the 101$^{st}$ sampling interval and keeps constant until the end of the cycle. It is observed that the proposed method dynamically yields the estimation at each sampling time and holds good accuracy.

To evaluate the overall estimation error over cycles, Fig. \ref{FIG11} presents the average estimation error at each sampling time over cycles. The estimation becomes reliable after the 51$^{st}$ sampling interval, from which the estimation accuracy becomes stable and close to the ground truth. In addition, the results are reasonable when compared with the sampling interval importance analysis given in Fig. \ref{FIG8}(a). Large estimation errors are observed before the 51$^{st}$ sampling interval due to the high data similarity over cycles. Due to the page limitation, the results from other batteries are not shown here, but similar results can be observed.

\section{Conclusion}
Inspired by the superiorities of the digital twin in describing the physical system with the massive offline data and online data, this work puts forward a digital twin model for Lithium-ion battery SOH estimation to accurately estimate maximum available capacity in a real-time way with only partial discharge data. This is the first implementation of such a critical topic through merging physical degradation behavior and advanced data-driven modeling. Through extensive results and comparisons, significant conclusions have been drawn. First, through the newly proposed energy discrepancy-aware time warping, the synchronized time series are consistent over cycles regarding temporal correlation, spatial shape, and energy discrepancy, avoiding overlooking or distorting any useful information. Second, in-depth process insights are gained with the synchronized cycling data, revealing the importance of different sampling times and facilitating the development of the time-attention SOH digital twin model. Moreover, real-time SOH estimation is gained with partially discharged data to estimate the capacity on the fly. Experiments on multiple battery cells demonstrate that the proposed digital twin provides excellent results for end-cycle and real-time estimations.

In addition to the above findings, the following promising research works deserve to further exploration:
\begin{enumerate}
  \item Model updating is necessary when online measurements are increasingly available. Transfer learning provides a rapid and efficient solution to address the distinction of data distribution between the incoming data and the historical data.
  \item The accurate detection of ``knee point'', where a battery begins to step into the degradation phase, is rarely discussed yet. The benefit is to avoid introducing slight degradation data into the real-time SOH estimation, which may weaken the estimation accuracy.
\end{enumerate}

\ifCLASSOPTIONcaptionsoff
\newpage
\fi
\bibliographystyle{IEEEtran}

\end{document}